\documentclass[10pt,twocolumn,letterpaper]{article}

\usepackage[pagenumbers]{cvpr} 

\definecolor{cvprblue}{rgb}{0.21,0.49,0.74}
\usepackage[pagebackref,breaklinks,colorlinks,allcolors=cvprblue]{hyperref}
\usepackage{algorithm}
\usepackage[T1]{fontenc}
\usepackage{lmodern}
\usepackage[noend]{algpseudocode}

\usepackage{multirow}
\usepackage{amssymb}
\usepackage{pifont}
\newcommand{\cmark}{\ding{51}}%
\newcommand{\xmark}{\ding{55}}%

\newcommand{\diff}[1]{#1}
\newcommand{\xg}[1]{\textcolor{blue}{#1}}
\newcommand{\ys}[1]{\textcolor{magenta}{#1}}

\def\paperID{4191} 
\def\confName{CVPR}
\def\confYear{2025}

\title{Alias-Free Latent Diffusion Models: \\ Improving Fractional Shift Equivariance of Diffusion Latent Space}

\author{Yifan Zhou$^1$\ \hspace{12pt} Zeqi Xiao$^1$\ \hspace{12pt} Shuai Yang $^{2}$\ \hspace{12pt} Xingang Pan$^{1}$\ \\
\normalsize{$^1 $S-Lab, Nanyang Technological University 
\hspace{12pt}
$^2$ Wangxuan Institute of Computer Technology, Peking University}\\
{\tt\small {\{yifan006, zeqi001, xingang.pan\}@ntu.edu.sg} \hspace{12pt} williamyang@pku.edu.cn}
}

\begin{document}

\twocolumn[{
    \renewcommand\twocolumn[1][]{#1}
    \maketitle
    \vspace{-10mm}
    \centering
    \includegraphics[width=0.9\linewidth]{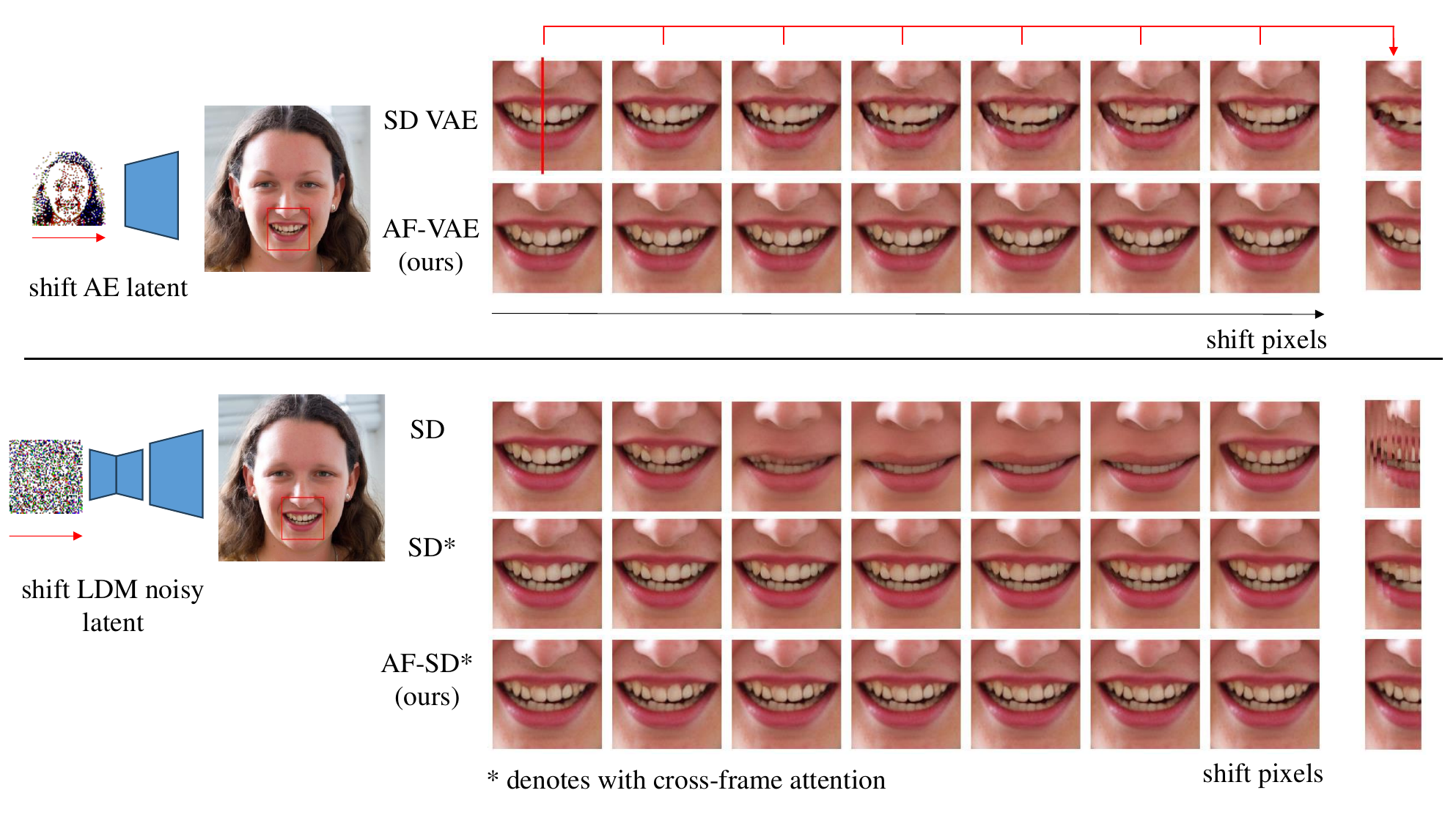}
    \vspace{-4mm}
    \captionof{figure}{Visualization of AE and LDM latent shift experiments. \textbf{Top}: we shift the input of an $8 \times$ upsampling VAE decoder by 1/8 pixel in each step. The intermediate output images from Stable Diffusion (SD) VAE appear increasingly blurry. \textbf{Bottom}: we shift the noisy latent of LDM by 1/8 pixel in each step. Without cross-frame attention, SD struggles to generate consistent results; even with it, we can observe a ``bouncing effect'', where textures initially stick but suddenly shift in subsequent steps. In contrast, our alias-free SD (AF-SD) produces shift-equivariant results in both experiments, maintaining consistency and clarity across steps. }
    \vspace{4mm}
    \label{fig:teaser}
}]

\maketitle

\begin{abstract}

Latent Diffusion Models (LDMs) are known to have an unstable generation process, where even small perturbations or shifts in the input noise can lead to significantly different outputs.
This hinders their applicability in applications requiring consistent results.
In this work, we redesign LDMs to enhance consistency by making them \textbf{shift-equivariant}.
While introducing anti-aliasing operations can partially improve shift-equivariance, significant aliasing and inconsistency persist due to the unique challenges in LDMs, including 1) aliasing amplification during VAE training and multiple U-Net inferences, and 2) self-attention modules that inherently lack shift-equivariance.
To address these issues, we redesign the attention modules to be shift-equivariant and propose an equivariance loss that effectively suppresses the frequency bandwidth of the features in the continuous domain. 
The resulting alias-free LDM (AF-LDM) achieves strong shift-equivariance and is also robust to irregular warping.
Extensive experiments demonstrate that AF-LDM produces significantly more consistent results than vanilla LDM across various applications, including video editing and image-to-image translation.
Code is available at: \url{https://github.com/SingleZombie/AFLDM}

\end{abstract}    
\section{Introduction}
\label{sec:intro}

Latent Diffusion Models (LDMs)~\cite{ldm} achieve high-resolution image synthesis by performing the denoising diffusion process in a compressed latent space obtained via a VAE~\cite{vae}.
Thanks to the rich image prior learned in LDMs, they have been broadly adopted in a variety of downstream applications.
For example, LDMs are widely used for video editing by processing each frame~\cite{text2video-zero, rerender, fresco, flatten, tokenflow, controlvideo, tune_a_video}.
They can also be finetuned for image-to-image translation tasks such as super-resolution~\cite{upscale, stablesr, sinsr} and normal estimation~\cite{stablenormal, geowizard, wonder3d, genpercept}, \etc
These tasks often require the results to be \textit{consistent}, \eg, the texture details of consecutive video frames should be coherent without flickering, and the normal estimation results of the same region should remain stable regardless of random shifts of the input image.

Despite the significant demand for ensuring such consistency, LDMs unfortunately fall short in this aspect due to the inherently unstable generation process from the Gaussian noise to the clean image.
Even small perturbations, such as pixel shifts or flow warping applied to the initial noisy latent, can result in drastically different images~\cite{rerender, warp_noise}, as shown in Fig.~\ref{fig:teaser} (SD and SD*).
While some attempts introduce additional sophisticated modules to alleviate the inconsistency~\cite{rerender}, how to fundamentally address this issue from the design of LDM remains underexplored.
This motivates us to investigate improving the \textit{shift-equivariance} of LDMs,  \ie, a shift to the latent noise (\eg, 1/8 pixel) should lead to a rescaled shift in the final generated image (\eg, 1 pixel), where the rescale is caused by the upsampling in the decoder.
This useful property can facilitate various vision synthesis tasks that require high stability and consistency.

Previous works have studied shift-equivariance for vanilla convolutional neural networks (CNNs) from the signal processing perspective~\cite{stylegan3, making_conv, af-convnet, aps}.
A common cause of the failure to preserve equivariance is \textit{aliasing effects}, which means that when we consider the discrete features in the continuous domain, they contain high frequencies beyond what the discrete sampling rate can represent according to the Nyquist–Shannon sampling theorem \cite{sampling_theorem}.
This happens in CNNs because some operations, such as downsampling, upsampling, and non-linear layers, cannot correctly band-limit the features in the continuous domain.
To address this issue, several anti-aliasing operations have been proposed, which constrain the output signal to be band-limited~\cite{stylegan3, making_conv, af-convnet}.
For example, StyleGAN3~\cite{stylegan3} has successfully adopted this principle to build an alias-free and shift-equivariant generator.

\begin{figure}[t]
\centering
\includegraphics[width=0.9\linewidth]{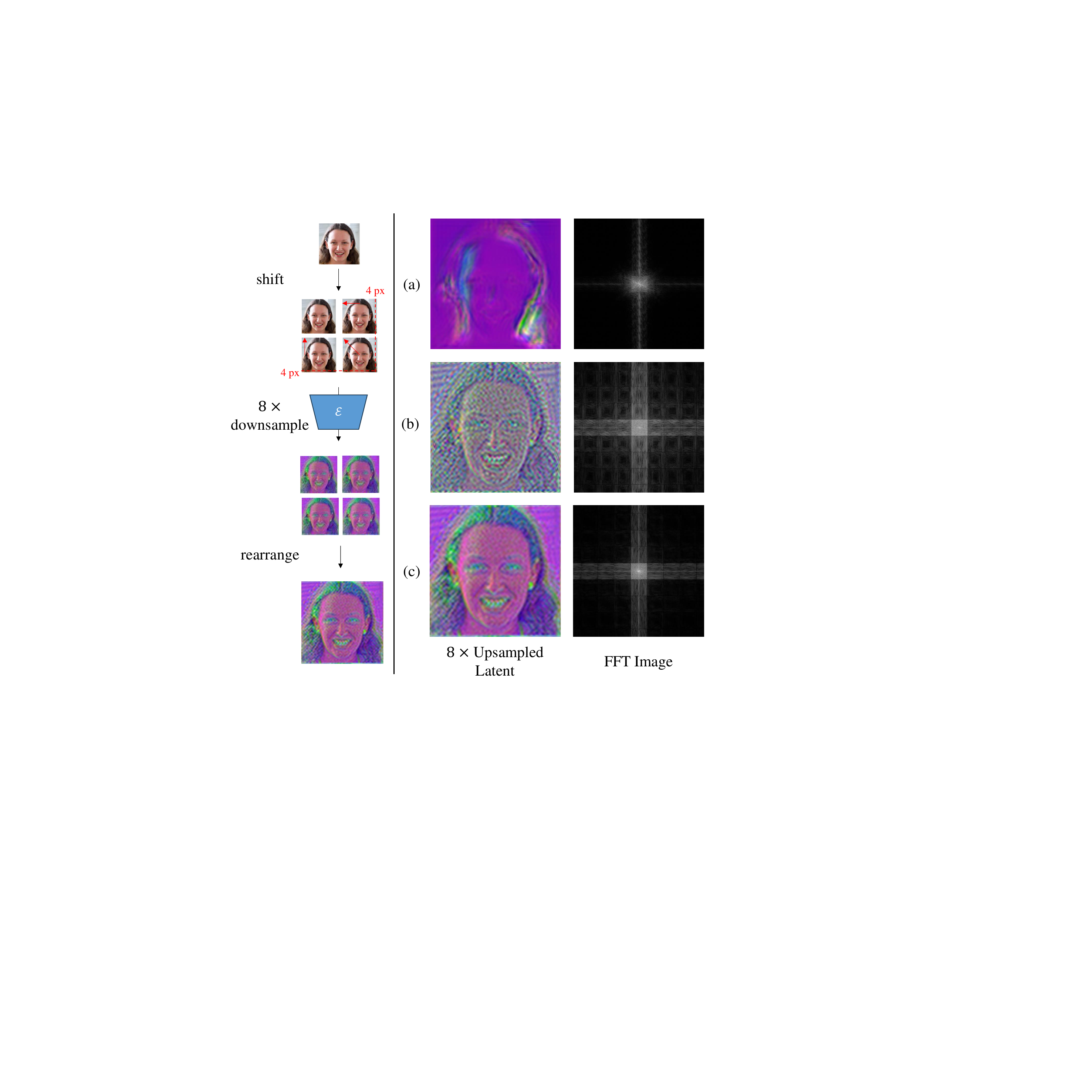}\vspace{-2mm}
\caption{Frequency map visualization of upsampled VAE latent. The upsampled latent serves as an approximation of a continuous signal. \textbf{Left}: Illustration of latent upsampling algorithm. For an encoder that downsamples the image by $8 \times$, we can leverage its shift-equivariance to achieve $2 \times$ upsampling of the latent. By shifting the input image by 4 pixels, we create a 0.5-pixel shift in the latent. Combining these fractional shifted latents yields a higher-resolution latent. \textbf{Right}: Visualization of $8 \times$ upsampled latent in both spatial and frequency domains. (a) AF-VAE with random weights, (b) AF-VAE without equivariance loss, (c) AF-VAE with equivariance loss. Ideally, a correctly band-limited latent would only use $1/8$ of the frequencies in the upsampled frequency domain. During training, the aliasing effects in a randomly initialized VAE with alias-free modules become more prominent; however, our proposed equivariance loss helps suppress the aliasing during training.}\vspace{-4mm}
\label{fig:fft_img}
\end{figure}

 A natural question then arises: Can we simply adopt those anti-aliasing modules in LDMs to achieve shift-equivariance? 
 Our preliminary study reveals that existing anti-aliasing modules alone are insufficient to achieve a highly shift-equivariant LDM.
 We identify several reasons: 
1) Although a randomly initialized VAE with alias-free modules initially exhibits reduced aliasing, these effects intensify as training progresses (Fig.~\ref{fig:fft_img}). 
This is likely because the VAE's learning process benefits from amplifying the remaining high-frequencies leaked from imperfect alias-free designs, such as the processing of boundary pixels~\cite{cnn_boundary} and nonlinearities \cite{af-convnet}.
2) The iterative denoising process in LDMs requires multiple U-Net inferences, causing aliasing to accumulate across denoising steps (Fig.~\ref{fig:denoise_diagram}).
3) The self-attention operations in the U-Net, which are sensitive to global translations, are not shift-equivariant.

To overcome these unique challenges in LDMs, we redesign alias-free LDMs in two distinct aspects. First, given the aliasing amplification in both VAE and U-Net inferences, it is extremely challenging to address aliasing purely from model design.
Instead, we introduce an effective but much less explored way, which is an equivariance loss that includes fractional shift-equivariance of LDM's VAE and U-Net as part of the training objective.
This significantly reduces aliasing without sacrificing performance.
Fig.~\ref{fig:fft_img} verifies that this loss effectively suppresses the bandwidth of the VAE latent space.

Second, we demonstrate that in order to make self-attention shift-invariant, the pool of keys and values must be fixed, while queries should shift along with the input. 
To achieve this, we select a reference frame, and always use the keys and values of that frame in attention calculation to ensure shift-equivariance with respect to that frame.
While this solution happens to have the same form as the cross-frame attention (CFA) \cite{text2video-zero}, we are the first to disclose its significance in equivariance.
In addition, while previous works often use CFA in test time~\cite{rerender, flatten, text2video-zero, fresco}, we find it critical to use CFA in both training and inference to suppress aliasing.

With these improvements, we present \textbf{Alias-Free Latent Diffusion Model (AF-LDM)}. 
As shown in Fig.~\ref{fig:teaser}, our AF-LDM demonstrates high fractional shift-equivariance throughout the generation pipeline. Further experiments verify that AF-LDM is robust even for irregular pixel shifts, such as flow warping. 
This property motivates us to propose a simple warping-equivariant video editing method that models the deformation among frames without using additional mechanisms for temporal coherence. 
Beyond generation, AF-LDM generalizes effectively to other image-to-image tasks that benefit from consistency and stability, such as super-resolution and normal estimation.

In summary, our contributions are as follows:

\begin{itemize}
	\item  We take the first attempt to investigate the shift-equivariance in LDMs and introduce a novel alias-free LDM (AF-LDM), which effectively mitigates aliasing effects and ensures shift-equivariance.
	
	\item  
	To address aliasing amplification in LDMs beyond the capabilities of existing techniques, we introduce an equivariance loss to constrain the bandwidth of the underlying continuous feature. 
	Additionally, we ensure shift-equivariance in attention layers by employing consistent reference keys and values during shifting or warping, applying this strategy in both training and inference stages.
	
	\item  We validate the effectiveness of AF-LDM in several downstream tasks that require consistent results, including video editing and image-to-image tasks, demonstrating its broad applicability and significant potential.

\end{itemize}

\section{Related Work}

\subsection{Diffusion Models}

The diffusion model is a family of generative models that transform Gaussian noise into a clean image through a progressive denoising process. Early diffusion models \cite{ddpm1, ddpm2} operate the denoising process directly in image space, requiring up to a thousand denoising steps. While effective, the image diffusion models are criticized for their extremely long training and inference time. Inspired by previous two-stage generative models \cite{vqvae, vqgan}, Latent Diffusion Models (LDMs)~\cite{ldm} accelerate the image diffusion models by decomposing the image generation process into latent generation followed by latent decoding. Although advancements in LDMs have introduced improvements from various perspectives, such as network backbone \cite{dit, sd3} and training objectives \cite{flow_match_1, flow_match_2, sit}, the two-stage generation structure remains integral. Our experiments reveal that all existing latent diffusion models suffer from instability caused by aliasing effects.

\subsection{Stability of Latent Diffusion Models}

Compared to GANs, LDMs exhibit notable instability, especially concerning smooth latent space and shift-equivariance. A smooth latent space, a concept well-developed in GAN literature, enables continuous changes in generated images when interpolating latents \cite{stylegan, stylegan2}. This smoothness enhances image editing applications for diffusion models, such as inversion \cite{ddim, null-text} and interpolation \cite{diffmorpher, dream_mover, attn_interp}. The other property, shift-equivariance, ensures consistency in generated outputs as input objects shift within a scene. This property is crucial for achieving temporal coherence in video editing applications~\cite{text2video-zero, rerender, fresco, flatten, tokenflow, warp_noise}. Common practices that mitigate the instability of diffusion models include image overfitting \cite{imagic, dragdiffusion, diffmorpher}, cross-frame attention \cite{text2video-zero}, and regularization \cite{smooth_diff, t_lipschitz}. Warped Diffusion \cite{warped_diffusion} specifically identifies the issue of shifting inconsistency in LDMs and proposes equivariance self-guidance as a remedy. Despite these efforts, the shift-equivariance of LDMs is still an underexplored topic.

\subsection{Alias-Free Neural Networks}

Aliasing, the overlap of signals due to improper sampling, significantly affects neural networks by reducing their shift-equivariance and shift-invariance \cite{making_conv, why_cnn}. Initial efforts to create alias-free neural networks focus on improving the shift-invariance of classification CNNs through optimized downsampling techniques \cite{making_conv, aps, tips}. Later, StyleGAN3 \cite{stylegan3} reveals that the upsampling, downsampling, and nonlinear operations of neural networks are not alias-free and introduce Kaiser filtering as an approximation to ideal signal processing. Following this, Alias-Free Convnets (AFC) \cite{af-convnet} expanded upon StyleGAN3's approach, replacing the Kaiser filters with FFT-based operations and implementing polynomial nonlinearities. While most previous works discuss the aliasing effects in classification CNNs and GANs, limited research has explored the aliasing effects in LDM.

\section{Method}
\label{sec:method}
\subsection{Preliminaries}
\label{sec:pre}
\textbf{Stable Diffusion}.
Stable Diffusion (SD)~\cite{ldm} is a latent diffusion model that consists of a Variational autoencoder (VAE) \cite{vae} and a text-conditioned denoising U-Net \cite{unet}. First, given an input image $x \in \mathbb{R}^{H \times W \times 3}$, the encoder $\mathcal{E}$ and decoder $\mathcal{D}$ of the VAE are trained to minimize the distance between $x$ and the reconstructed image $\mathcal{D}(z)$, where $z = \mathcal{E}(x) \in \mathbb{R}^{H/k \times W/k \times 4}$ is the latent downsampled by $k \times$. Next, a latent denoising U-Net $\epsilon_{\theta}(z_t, t)$ is trained to denoise the noisy latent $z_t$ at timestep $t$. During sampling, a clean latent $z$ is sampled via a reverse diffusion process and then decoded to obtain the final image $\mathcal{D}(z)$.

Both VAE and U-Net are built on the encoder-decoder architecture that consists of similar modules. Specifically, the networks incorporate ResNet blocks \cite{resnet} and self-attention blocks \cite{transformer}. To rescale the feature, the networks employ the nearest downsample and bilinear upsample. Although the networks heavily rely on shift-equivariant convolution layers, the other modules, including nonlinearities, sampling, and self-attention, are not shift-equivariant.

\noindent
\textbf{StyleGAN3}
StyleGANs \cite{stylegan, stylegan2} generate high-resolution images by gradually upsampling a constant low-resolution feature with a CNN-based synthesis network. Despite high generation quality, a ``texture sticking" effect occurs in the original StyleGAN, where the details in the output image stick to the same position when shifting the input feature. StyleGAN3 \cite{stylegan3} argues that the aliasing in the synthesis network causes this effect. The issue is analyzed by a continuous signal interpretation: each network feature is equivalent to a continuous signal. A discrete feature $Z$ can be converted into a continuous signal $z$ via interpolation filter and be converted back via Dirac comb. Similarly, each network layer $F$ has its counterpart $f$ in the continuous domain. If a network layer is shift-equivariant, then its effect in the discrete and continuous domain should be equivalent, \textit{i.e.}, the conversion between $F(Z)$ and $f(z)$ should still be invertible. 


Based on the continuous signal interpretation, StyleGAN3 makes several primary improvements:

1. Fourier Latent: Transform the input of the synthesis network from the image feature to the Fourier feature.

2. Cropping Boundary Pixels: Crop the border pixels after each layer to stop leaking absolute position.

3. Ideal Resampling: Replace bilinear upsampling with Kaiser filters that better approximate an ideal low-pass filter.

4. Filtered Nonlinearity: Wrap nonlinearities between a $2 \times$ upsampling and $2\times$ downsampling to suppress the high frequencies introduced by this operation.

\noindent
\textbf{Fractional Shift Equivariance}.
Let $T_{\Delta}$ be the shift operator that shifts an image $x \in \mathbb{R}^{H \times W}$ by $\Delta \in \mathbb{N}^2$ pixels, an operator $f: \mathbb{R}^{H \times W} \to \mathbb{R}^{H \times W}$ is shift equivariant if it satisfies $ f(T_{\Delta}(x)) = T_{\Delta}(f(x))$. The shift equivariance can be extended to fractional shift equivariance, \textit{i.e.}, $\Delta \in \mathbb{R}^2$. 
In this work, we implement it as the Fourier shift, which shifts the phase of an image in the Fourier domain according to $\Delta$. With fractional shift, we can extend $f$ to an operator that rescales the image. Assume $f: \mathbb{R}^{H \times W} \to \mathbb{R}^{kH \times kW}$ is an operator that rescale the image by a factor of $k$, then it is fractional shift equivariant if $ f(T_{\Delta}(x)) = T_{k \cdot \Delta}(f(x))$.

Since the resolution of an image is limited, some pixels are moved out or missing after shifting. There are two common ways to process the edge pixels: circular shift $T^{\text{cir}}$ fills the missing pixels with pixels that just moved out. In constant, cropped shift $T^{\text{cro}}$ fills the missing pixels with a constant, and the shift-equivariance is only considered in valid regions. This paper uses cropped shift by default since it is closer to the default padding mode of convolutional layers.

\subsection{Alias-free Latent Diffusion Models}

\begin{figure}[t]
\centering
    \vspace{-2mm}
    \includegraphics[width=\linewidth]{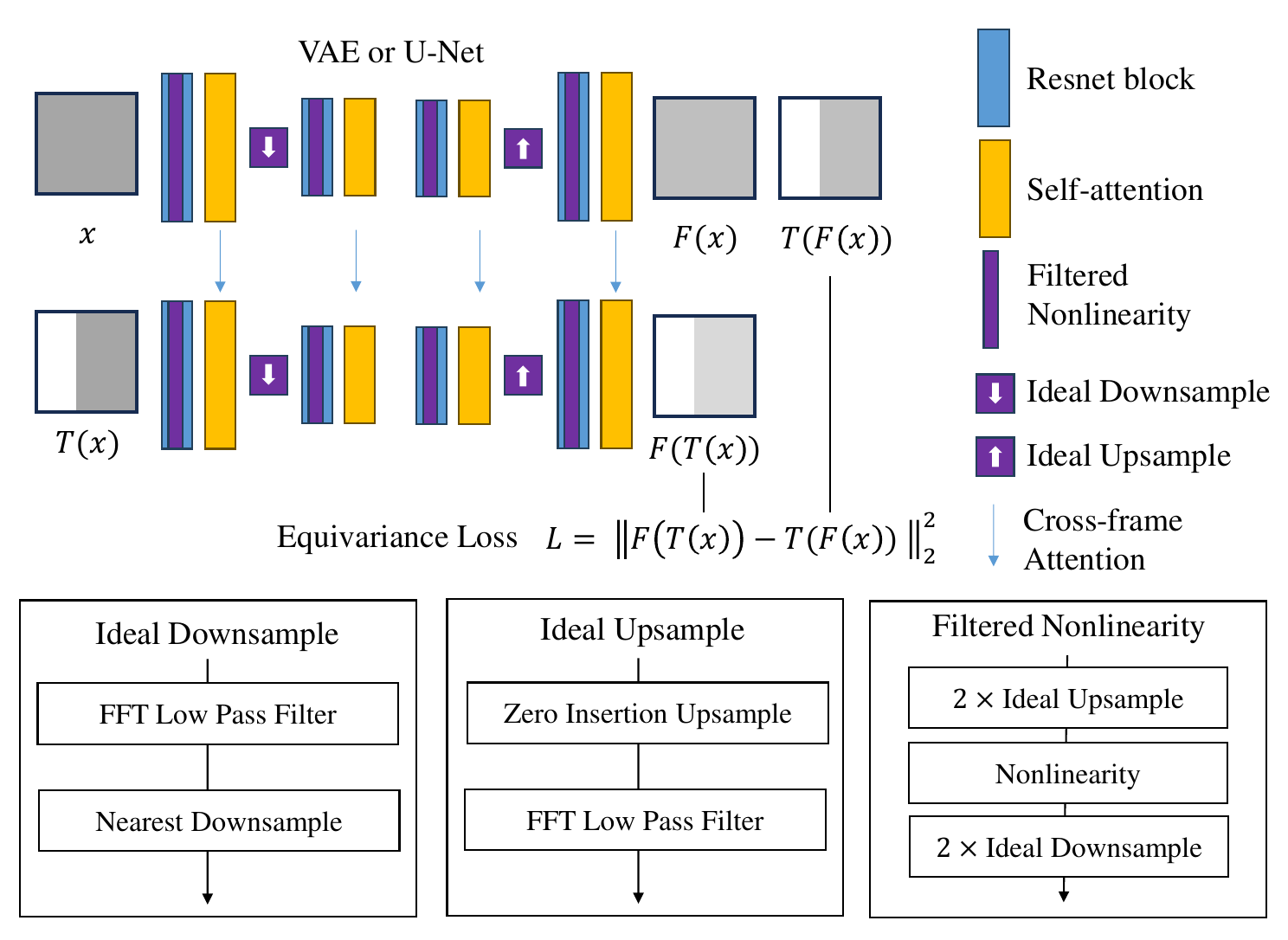}
    \vspace{-6mm}
    \captionof{figure}{The architecture of AF-LDM. Both VAE and U-Net of SD can be represented by an encoder-decoder structure. We implement ideal upsample, ideal downsample and filtered nonlinearity following \cite{stylegan3, af-convnet}. }
    \vspace{-4mm}
    \label{fig:arch}
\end{figure}

In this work, we focus on improving the fractional shift-equivariance of SD~\cite{ldm} for its widespread applications in the community. While StyleGAN3 provides an effective methodology for designing alias-free CNNs, these alias-free mechanisms cannot be directly applied to SD. This is because the generation process of StyleGAN3 solely depends on the latent-to-image synthesis network, enabling extremely flexible latent-related design for equivariance, such as using Fourier latents and cropping border pixels as reviewed in Sec.~\ref{sec:pre}.
In contrast, the equivariance of LDMs depends jointly on AE encoder, decoder, and U-Net. Specifically, the latent is produced by AE encoder, leaving less room to redefine the latent, not to mention the self-attention operators that are sensitive to small changes. 



To this end, we propose Alias-Free LDM (AF-LDM) (Fig.~\ref{fig:arch}), which has two significant improvements. 1) We propose a continuous latent representation (Sec.~\ref{sec:clr}), which has a limited frequency bandwidth and naturally supports fractional shift. To ensure this representation, we modify network modules with anti-aliasing designs and regularize the learning process with an equivariance loss. 2) To enhance the equivariance of self-attention, we employ equivariant attention (Sec.~\ref{sec:cfa}) that fixes the key and value features of self-attention when processing the shifted latents. Some anti-aliasing techniques are inspired by StyleGAN3. The comparison between StyleGAN3 and our AF-LDM is summarized in Table~\ref{tab:vs_stylegan3}. 

\begin{table} []
\caption{Comparison between StyleGAN3 and AF-LDM. Our main improvements are highlighted in bold. }\vspace{-2mm}
\label{tab:vs_stylegan3}
\resizebox{\linewidth}{!}{
\centering
\begin{tabular}{l|c|c}
\toprule
 & StyleGAN3 & AF-LDM  \\
\midrule 
Latent Representation & Fourier feature & Continuous latent \\
Resampling & Kaiser filtering & FFT \\
Filtered Nonlinearity & \cmark & \cmark \\
Cropping Boundary Pixels & \cmark & \xmark \\
\textbf{Equivariance Loss} & \xmark & \cmark \\
\textbf{Equivariant Attention} & - & \cmark \\
\bottomrule
\end{tabular}}\vspace{-4mm}
\end{table}

\subsection{Continuous Latent Representation via Equivariance Loss}
\label{sec:clr}
Based on the definition, a LDM pipeline is fractional shift-equivariant if 

\begin{equation}
    \epsilon_{\theta}(T_{\Delta}(z_t), t) = T_{\Delta}(\epsilon_{\theta}(z_t, t)), 
\end{equation}
\begin{equation}
    \mathcal{D}(T_{\Delta}(z)) = T_{k \cdot \Delta}(\mathcal{D}(z)), 
\end{equation}
where $k$ is the downsampling factor of VAE. Since we implement fractional shift as Fourier shift, each $z$ should be a band-limited continuous latent that can be sampled everywhere by applying DFT, phase shift, and IDFT. For example, if the resolution of input image $x$ is $s \times s$, then the maximum frequency of $z$ should be less than $s / (2k)$ based on sampling theorem~\cite{sampling_theorem}.


By incorporating anti-aliasing layers shown in Fig.~\ref{fig:arch}, the shift-equivariance of a randomly initialized AE is significantly enhanced\footnote{We found the effect of self-attention layers in VAE is neglectable. We do not apply Equivariant Attention to VAE.}, as verified in Table~\ref{tab:vae}. However, as training progresses, shift-equivariance gradually deteriorates. We hypothesize that this is because of the imperfect anti-aliasing designs, such as the processing of boundary pixels and nonlinearities, which enable the network to introduce aliasing that undermines equivariance. These aliasing effects may inadvertently aid the networks' learning process, leading to their amplification over time.


Recognizing that anti-aliasing modules alone are insufficient to fully resolve the equivariance issue, we shift our focus to regularizing network learning by directly optimizing the error between the outputs of shifted inputs and the corresponding shifted outputs. Specifically, we propose an equivariance loss for any network $f$ that rescales the input by $k \times $:
\begin{equation}
    L = || f(T_{\Delta}(x)) - T_{k \cdot \Delta}(f(x)) ||_2^2.
\end{equation}
The loss is directly added to the original training loss of $f$. Since $T$ is a Fourier shift, this loss forces the network to only use the low-frequency information that the discrete latent can represent. With all these modifications, the continuous latent of Alias-free VAE is nearly a band-limit signal as illustrated in Fig.~\ref{fig:fft_img}. In practice, the equivariance loss is implemented as follows.

\noindent \textbf{VAE}.
We compute the equivariance loss for the VAE encoder and decoder separately.  Formally, given input $x \in \mathbb{R}^{H \times W \times C}$, the equivariance loss is defined as

\begin{equation}
    L_{\text{enc}} = || \left[ \mathcal{E}(T_{\Delta}(x)) - T_{\Delta / k}(\mathcal{E}(x)) \right] \cdot M_{\Delta / k} ||_2^2,
\end{equation}
\begin{equation}
    L_{\text{dec}} = || \left[ \mathcal{D}(T_{\Delta / k}(z)) - T_{\Delta  }(\mathcal{D}(z)) \right] \cdot M_{\Delta} ||_2^2,
\end{equation}
where $z = sg(\mathcal{E}(x))$, $sg$ is stop gradient operator, and $M_{\Delta}$ denotes the valid mask for cropped shift $T_{\Delta}$. Since the fractional shift of an image is not well defined, the offsets are integers, \textit{i.e.}, $\Delta=(\Delta_x, \Delta_y) \in \mathbb{N}^2$. 


\noindent \textbf{U-Net}. For latent $z_t \in \mathbb{R}^{H/k \times W/k \times C'}$ at timestep $t$, the equivariance loss of U-Net is defined as

\begin{equation}
    L_{\text{unet}} = || \left[ \epsilon'_{\theta}(T_{\Delta}^{\text{cir}}(z_t), t) - T_{\Delta}(\epsilon_{\theta}(z_t, t)) \right] \cdot M_{\Delta} ||_2^2,
\end{equation}
where $\Delta=(\Delta_x / k, \Delta_y / k)$, $\Delta_x$ and $\Delta_y$ are sampled in the same way as VAE. $\epsilon'_{\theta}$ is the U-Net with equivariant attention (Sec.~\ref{sec:cfa}). We first compute $\epsilon_{\theta}(z_t, t)$ and cache attention features and then compute $\epsilon'_{\theta}(T_{\Delta}^{\text{cir}}(z_t), t)$. Considering U-Net is more easily affected by the padded pixels due to cropped shift in the iterative denoising process, we apply circular shift $T^{\text{cir}}$ when shifting the input latent to pad meaningful pixels. Note that we still use cropped valid mask $M_{\Delta}$ to filter out the padded pixels when computing loss.

\subsection{Equivariant Attention}
\label{sec:cfa}


Given an input image token sequence $x \in \mathbb{R}^{HW \times d_m}$, the self-attention operation \cite{transformer} is defined as 
\begin{equation}
    \text{SA}(x) = \text{softmax}(xW^Q(xW^K)^\top)xW^V,
\end{equation}
where $W^Q, W^K \in \mathbb{R}^{d_m \times d_k}$ and $W^V \in \mathbb{R}^{d_m \times d_v}$ are projection matrices. For simplicity, the scaling factor is omitted. \diff{While self-attention is inherently shift-equivariant under circular shifts, it fails to maintain equivariance under cropped shifts or non-rigid deformations (e.g., warping). That is because to achieve equivarince, we need} 

\begin{equation}
\label{equ:sa_2}
\begin{split}
    \text{softmax}(T(x)W^Q(T(x)W^K)^\top)T(x)W^V = \\ 
T(\text{softmax}(xW^Q(xW^K)^\top)xW^V),
\end{split}
\end{equation}
\diff{and $T(\cdot)$ denotes shifting (reindexing rows). The right-hand side of the equation~\ref{equ:sa_2} reduces to $\text{softmax}(T(x)W^Q(xW^K)^\top)xW^V$ since softmax and matrix right multiplication are row-wise independent. The equation breaks since $T(x)W^K$ and $T(x)W^V$ are modified by cropping.}

To address this, we redefine self-attention as a point-wise operation, \diff{ensuring equivariance under any \textit{relative} shifts to a reference frame}. Formally, given a reference \diff{frame} $x_r \in \mathbb{R}^{HW \times d_m}$ and its shifted version $x_s$, we propose Equivariant Attention (EA), defined as:

\begin{equation}
    \text{EA}(x_r, x_s) = \text{softmax}(x_sW^Q(x_rW^K)^\top)x_rW^V.
\end{equation}
Since matrix right multiplication is row-wise independent \diff{and each token is a row vector of $x_s$}, the above equation is a point-wise operation applied individually to each token in $x_s$. As a result, the operation is inherently equivariant for any \textit{relative} deformation. 

Interestingly, we observe that this equivariant attention mechanism has the same form as Cross-Frame Attention (CFA) in video editing literature, which is primarily used as an inference-time technique.
In our method, however, it is important to incorporate EA together with equivariance loss during training to effectively mitigate aliasing effects.
This is because, without EA, the equivariance loss will focus on fixing the attention module rather than the aliasing.
For ease of understanding, we refer to EA as CFA in the experiment section. 

\section{Experiment}

\subsection{Ablation Study}

\begin{table} []

\caption{AF-VAE reconstruction quality and shift-equivariance on $256 \times 256$ ImageNet validation set. Equiv. Loss refers to Equivariance Loss. $\dagger$: Original SD VAE was trained on OpenImages \cite{openimages}. This model is trained on ImageNet using our code. }\vspace{-2mm}
\label{tab:vae}
\resizebox{\linewidth}{!}{
\centering
\begin{tabular}{l|c|c|c|c}
\toprule
Model Configuration & Rec. PSNR & rFID & Enc. SPSNR & Dec. SPSNR  \\
\midrule 
SD VAE              & \textbf{26.94} & 0.69 & 25.07 & 22.31 \\
SD VAE$\dagger$     & 26.45 & \textbf{0.63} & 19.71 & 21.36 \\
+ Ideal sampling    & 25.76 & 1.01 & 21.51 & 21.75 \\
+ Filtered nonlinearity & 25.42 & 0.86 & 29.49 & 24.61 \\
+ Equiv. loss (AF-VAE)   & 25.53 & 1.00 & \textbf{45.10} & \textbf{33.88} \\
\midrule
SD VAE$\dagger$ + Equiv. loss & 24.59 & 4.18 & 41.46 & 30.74 \\
AF-VAE random weights & - & - & 29.43 & 29.02 \\

\bottomrule
\end{tabular}}\vspace{-2mm}
\end{table}

\begin{table} []
\caption{Unconditional AF-LDM sampling quality and shift-equivariance on $256 \times 256$ FFHQ. }\vspace{-2mm}
\label{tab:ldm}
\resizebox{\linewidth}{!}{
\centering
\begin{tabular}{l|c|c|c}
\toprule
Model Configuration & FID & Latent SPSNR & Image SPSNR  \\
\midrule 
LDM w/ SD VAE           & 14.91 & 27.47 & 12.38  \\
+ CFA in Inference         & 14.91 & 38.22 & 21.69  \\
\midrule
LDM w/ AF-VAE           & 15.27 & 22.87 & 12.18 \\
+ CFA in Inference      & 15.27 & 23.58 & 13.09 \\
+ Ideal Sampling        & 17.03 & 24.44 & 14.03 \\
+ Filtered Nonlinearity & \textbf{14.28} & 24.53 & 14.15 \\
+ Equiv. Loss w/o CFA    & 16.90 & 28.57 & 17.65 \\
+ CFA in Equiv. Loss (AF-LDM) & 19.05 & \textbf{40.94} & 28.06\\
\midrule
LDM w/ AF-VAE + Equiv. loss  & 39.14 & 37.63 & 25.99\\
AF-LDM random weights       & -     & 39.84 & \textbf{29.68}\\

\bottomrule
\end{tabular}}\vspace{-2mm}
\end{table}

\begin{figure}[t]
\centering
\includegraphics[width=1.0\linewidth]{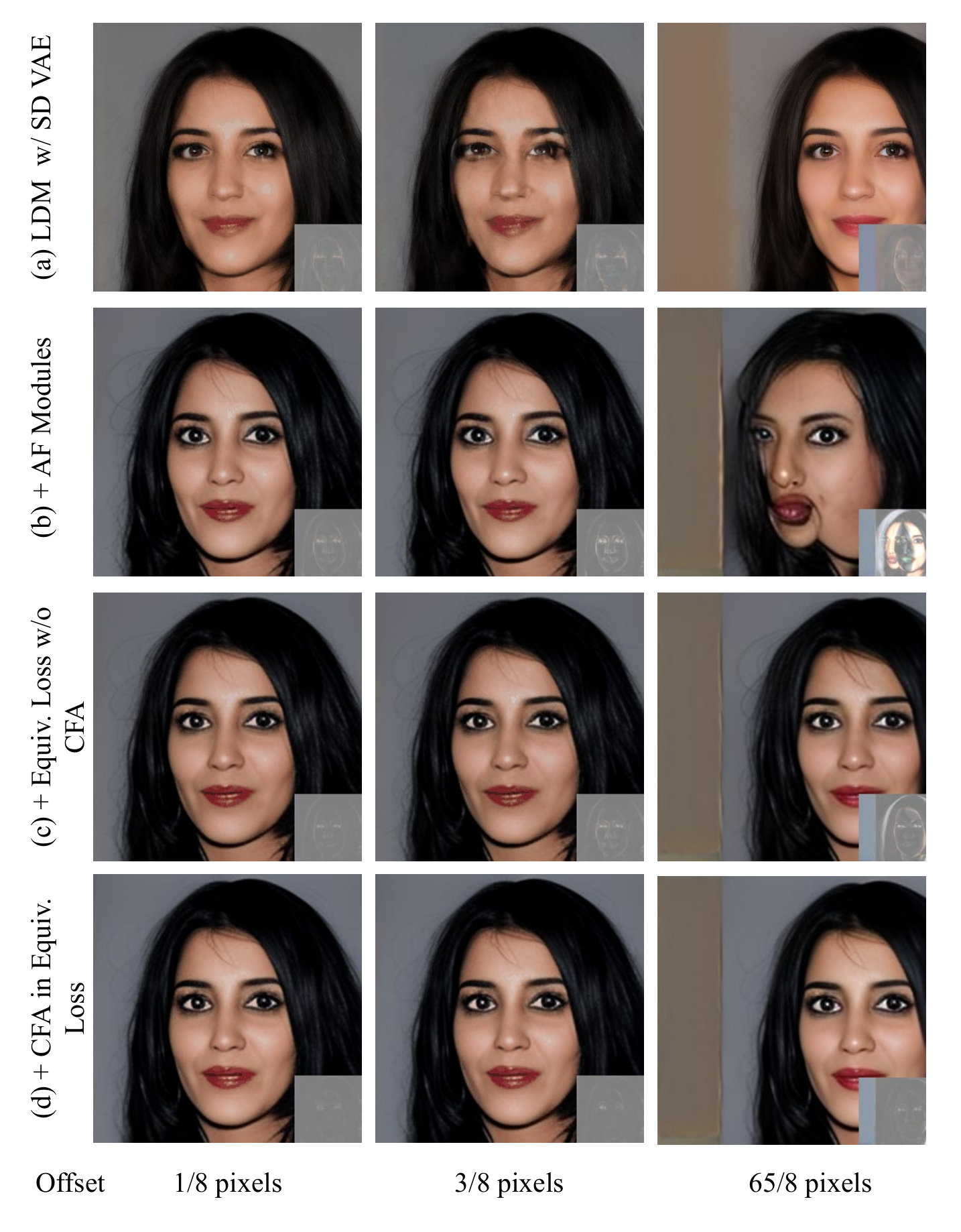}\vspace{-2mm}
\caption{Visualization of LDM latent fractional shifts. The latent is obtained by DDIM inversion.  The difference map between outputs and shifted outputs is given in the bottom right corner.}\vspace{-3mm}
\label{fig:ldm_shift}
\end{figure}


In the ablation study, we first train the Alias-Free VAE (AF-VAE) on $256 \times 256$ ImageNet \cite{imagenet}, initializing weights from the Stable Diffusion VAE (kl-f8 model in \cite{ldm}). We then train an unconditional AF-LDM in the latent space of AF-VAE on $256 \times 256$ FFHQ \cite{stylegan} from scratch. 

To assess the fractional shift-equivariance of models, we compute shift PSNR (SPSNR) \cite{making_conv, stylegan3} that positively correlates with shift-equivariance. Formally, for a model $f$ that rescales the inputs by $k$, SPSNR for an input $x$ is given by:
\begin{equation}
    \text{SPSNR}_f(x) = \text{PSNR}(f(T_{\Delta}(x)), T_{k \cdot \Delta}(f(x))).
\end{equation}

In the following experiments, we evaluate both the quality and shift-equivariance of VAE and LDM. Our alias-free models are trained with an objective that balances enhancing equivariance with maintaining the original task quality.

\noindent \textbf{VAE}. To evaluate VAE quality, we use reconstruction PSNR and FID \cite{fid}. Shift-equivariance is measured by encoder SPSNR and decoder SPSNR. The encoder SPSNR checks if the VAE latents are properly band-limited, while the decoder SPSNR evaluates output consistency under input shift. All metrics are computed on 50,000 $256 \times 256$ ImageNet validation images. 


Table \ref{tab:vae} shows that ideal sampling and filtered nonlinearity improve VAE shift-equivariance. However, both encoder and decoder SPSNR decrease compared to a randomly initialized model (last row), possibly due to imperfect anti-aliasing modules. Adding an equivariance loss as a regularization term balances task fidelity with high shift-equivariance. Note that applying equivariance loss without anti-aliasing modules significantly lowers the reconstruction quality (second-to-last row), highlighting that alias-free architecture is necessary for adding equivariance loss.

\noindent \textbf{LDM}. For LDM quality evaluation, we use FID computed on 50,000 samples generated with 50 DDIM steps. Shift-equivariance of LDM is measured by latent SPSNR and image SPSNR, with the former only considering the denoising process and the latter encompassing both denoising and VAE encoding. The SPSNR is computed on 10,000 randomly sampled noisy latents. Unless otherwise specified, CFA is enabled by default during inference, as disabling it would result in significant changes to the image's identity.


The quantitative results are shown in Table \ref{tab:ldm}, and partial visualization results are shown in Fig.~\ref{fig:ldm_shift}. Although LDM with SD VAE has good SPSNR, we can observe severe quality degradation in fractional shift results (Fig.~\ref{fig:ldm_shift}(a)). Using AF-VAE can improve the quality of shifted frames, but ``texture sticking" still occurs (Fig.~\ref{fig:ldm_shift}(b)). This issue can be addressed by adding equivariance loss (Fig.~\ref{fig:ldm_shift}(c)). By applying CFA in equivariance loss, the final model, AF-LDM, has the best overall shift-equivariance.

\begin{figure}[t]
\centering
\includegraphics[width=0.6\linewidth]{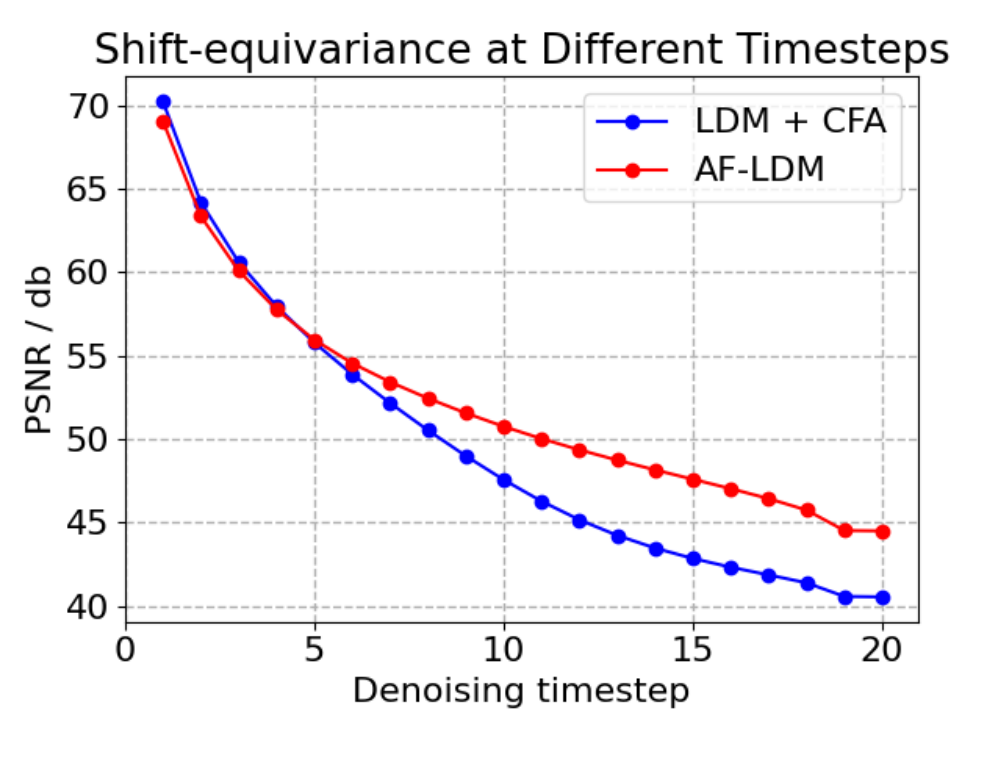}\vspace{-2mm}
\caption{Shift-equivariance during denoising process. We perform two denoising processes: one with Gaussian noise and the other with a $1/2$ pixel-shifted version of the same noise, computing the SPSNR after each denoising step. The experiment is conducted on 10,000 samples over 20 denoising steps. }\vspace{-4mm}
\label{fig:denoise_diagram}
\end{figure}

To further demonstrate the improvements of our AF-LDM over standard LDM, we evaluate SPSNR at each step of the denoising process, as shown in Fig.~\ref{fig:denoise_diagram}. In standard LDM, shift-equivariance degrades progressively due to error accumulation throughout the denoising process. In contrast, our AF-LDM mitigates this degradation, maintaining higher shift-equivariance across denoising steps.

\begin{figure}[t]
\centering
\includegraphics[width=1.0\linewidth]{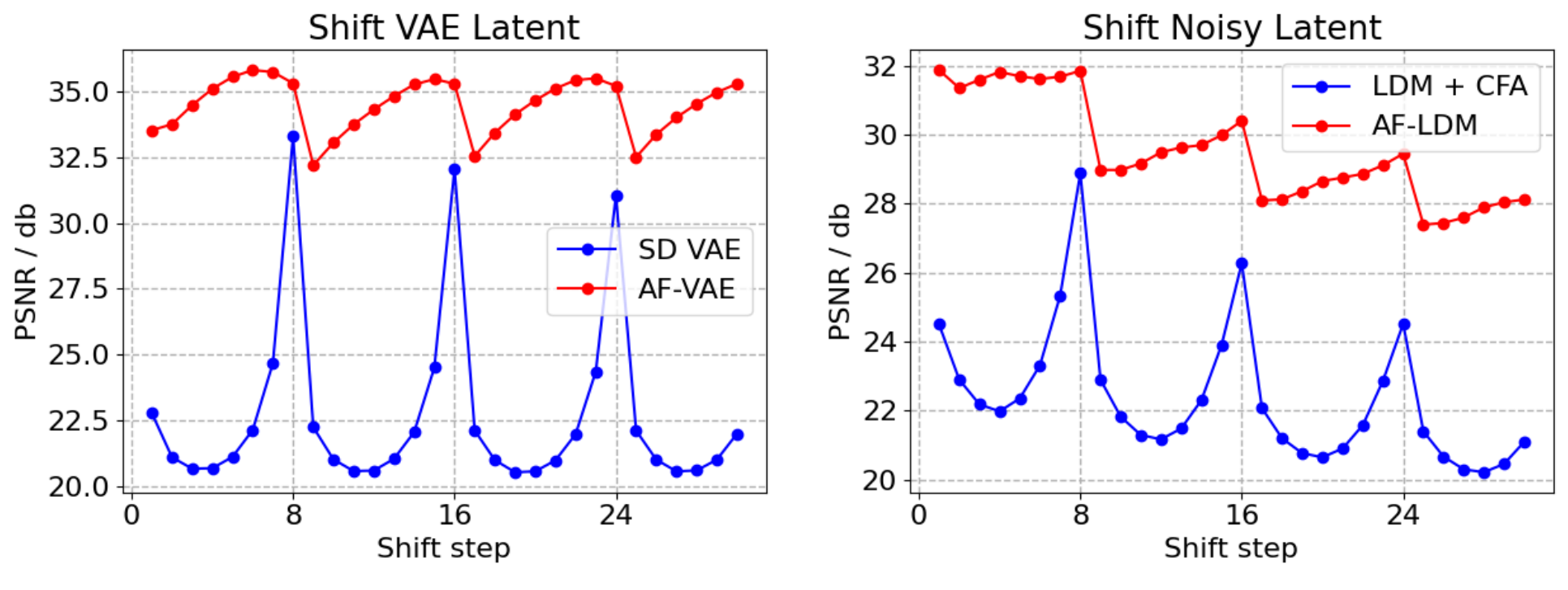}\vspace{-2mm}
\caption{Quantitative comparison of fractional shift consistency. We horizontally shift latents by $1/8$ pixels per step and compute the average SPSNR for the VAE decoder and LDM pipeline for each step. The VAE is tested on 500 ImageNet validation images, and LDM is tested on 500 FFHQ images.}\vspace{-2mm}
\label{fig:shift_diagram}
\end{figure}

\noindent \textbf{Fractional Shift Consistency}. To verify consistent shift-equivariance under fractional shifts, we horizontally shift a set of images and compute the averaged SPSNR for each step (Fig.~\ref{fig:shift_diagram}). Baseline models exhibit high shift-equivariance only for integer shifts, causing the blur outputs of the VAE decoder and the ``bouncing effect'' of LDM. In contrast, alias-free models achieve higher average SPSNR and lower variance, resulting in more stable AF-LDM outputs.

\subsection{Warping-Equivariant Video Editing}

\begin{figure}[t]
\centering
\includegraphics[width=1.0\linewidth]{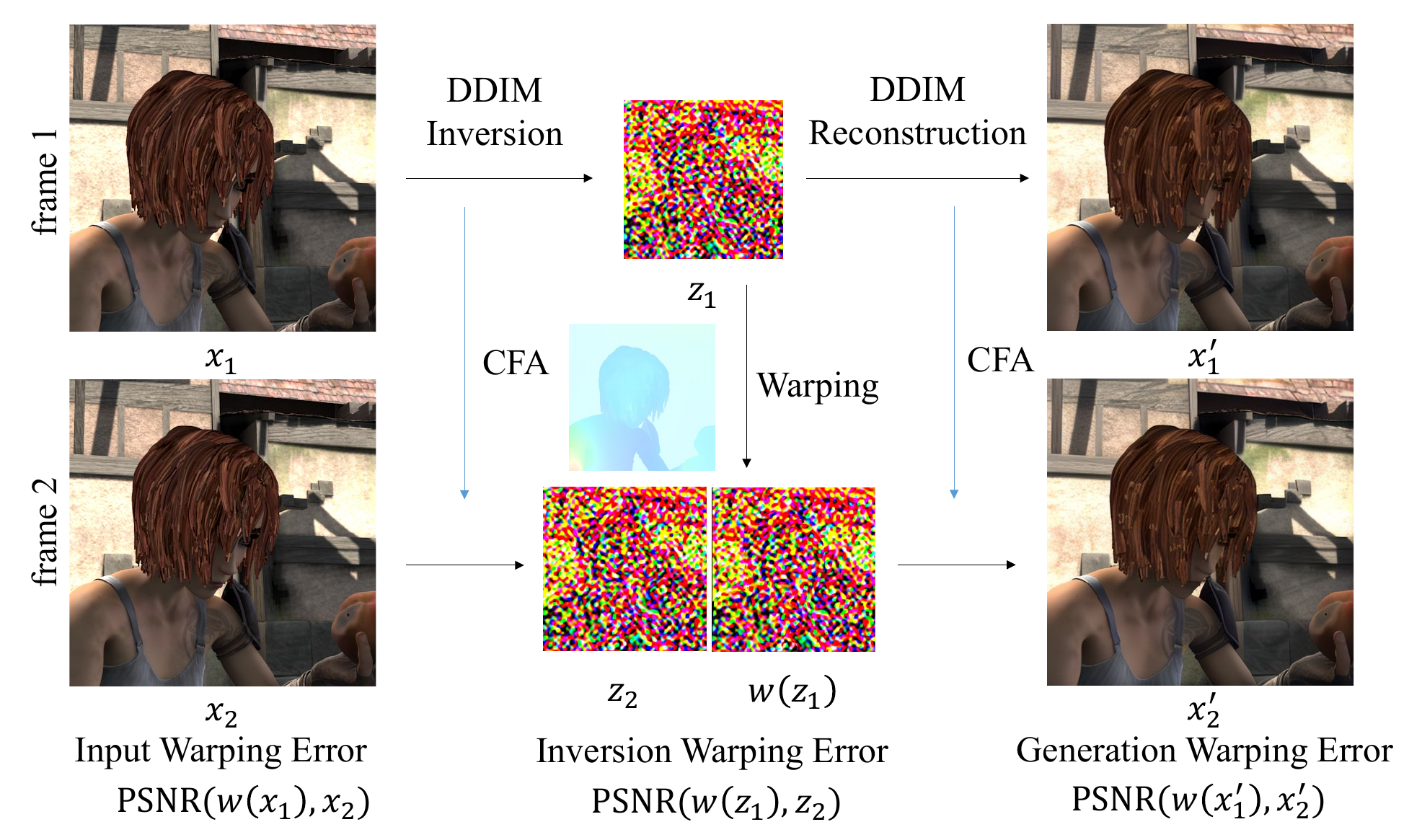}\vspace{-2mm}
\caption{Warping consistency experiments. Given neighboring frames and their ground truth optical flow, we can warp the inverted latents and compute the inversion warping error. We then reconstruct an inverted latent and its warped version to test the generation warping error. Cross-Frame attention (CFA) is enabled in both inversion and generation. The input warping error is computed as a reference. }\vspace{-4mm}
\label{fig:warping_error}
\end{figure}

With a more stable generation process, our AF-LDM can enhance video editing by improving temporal consistency. To showcase this application, we implement an alias-free Stable Diffusion (AF-SD) by integrating AF-VAE, anti-alias U-Net modifications, and retraining the text-conditional LDM on LAION Aesthetic 6.5+ dataset~\cite{laion}. We then propose a simple and elegant \textit{Warping-equivariant} video editing algorithm.

\begin{figure}[t]
\centering
\includegraphics[width=0.9\linewidth]{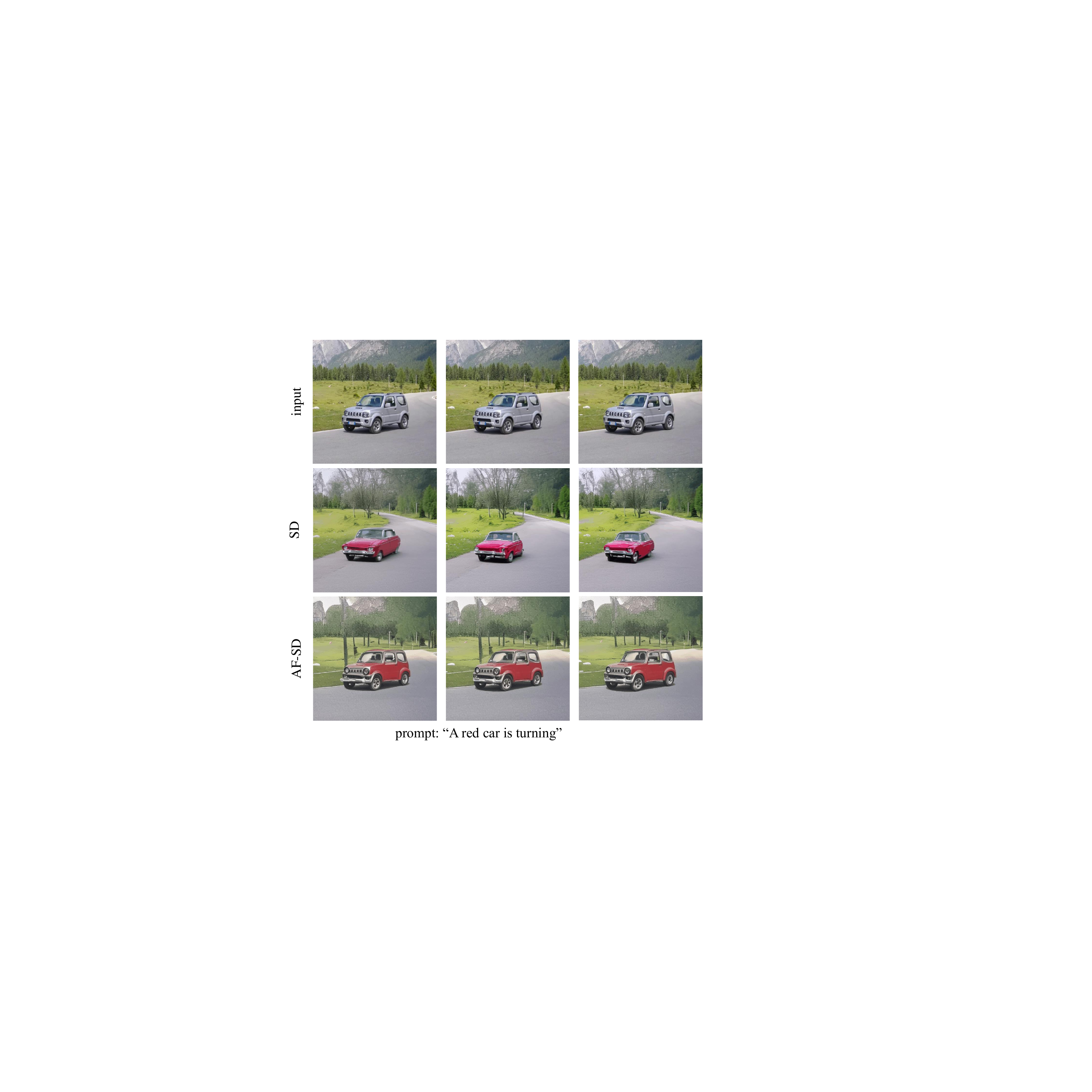}\vspace{-2mm}
\caption{Qualitative comparison of video editing consistency between SD and AF-SD using warping-equivariant video editing. Texture flickering can be observed in SD's results. More results are available in the supplementary. }\vspace{-4mm}
\label{fig:video_editing}
\end{figure}

Although our AF-SD is designed for pixel shift-equivariance, it performs well under irregular pixel shifts, such as flow warping. We verify this through the experiment shown in Fig.~\ref{fig:warping_error}.
Quantitative results in Table~\ref{tab:sd} demonstrate superior inversion and generation warping-equivariance in AF-SD over SD, indicating that AF-SD can better preserve deformation information. Higher inversion warping-equivariance is especially advantageous in inversion-based video editing because even if the reconstruction process is changed (for example, using another prompt), the diffusion model can still generate consistent results by utilizing the deformation information from latents. 

Based on these observations, we propose a novel warping-equivariant video editing method. First, we apply DDIM inversion to input video to an intermediate denoising timestep (similar to SDEdit \cite{sdedit}) with an empty prompt. Then, we regenerate a video from the inverted latents with a new prompt. CFA is enabled in both processes. This method implicitly deforms noisy latents without the need for explicit latent warping \cite{rerender, fresco, flatten, warp_noise}. As illustrated in Fig.~\ref{fig:video_editing}, AF-SD's robust inversion and generation ensure more consistent edited videos than SD.


\begin{table}
\caption{Warping error (defined in Fig.~\ref{fig:warping_error}) computed on Sintel~\cite{sintel}. }\vspace{-2mm}
\label{tab:sd}
\resizebox{\linewidth}{!}{
\centering
\begin{tabular}{l|c|c|c}
\toprule
Model  & Input Warping PSNR & Inversion Warping PSNR & Generation Warping PSNR  \\
\midrule 

SD  & \multirow{2}*{22.77} & 17.53 & 21.95 \\
AF-SD  &  & 19.68 & 26.10  \\

\bottomrule
\end{tabular}}\vspace{-3mm}
\end{table}



\subsection{Image-to-image Translation}

\begin{figure}[t]
\centering
\includegraphics[width=1.0\linewidth]{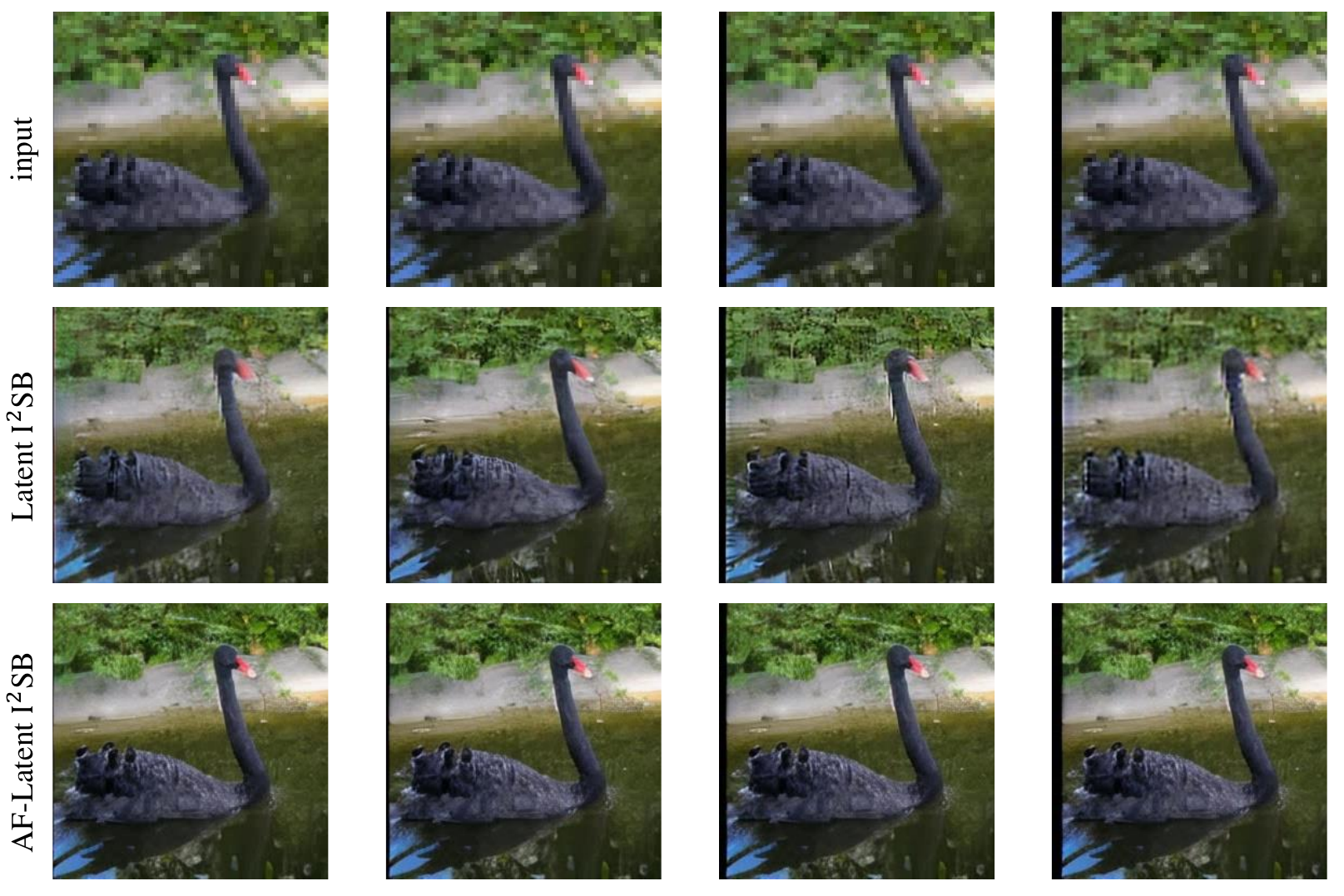}\vspace{-3mm}
\caption{Qualitative comparison of shift-equivairance in latent I$^2$SB $4 \times$ super-resolution results.}\vspace{-4mm}
\label{fig:i2sb}
\end{figure}

\noindent \textbf{Super-resolution with Latent I$^2$SB}. Image-to-Image Schr{\"o}dinger Bridge (I$^2$SB) \cite{i2sb} is a conditional diffusion model that directly maps one image distribution to another without going through Gaussian noise. Originally, I$^2$SB operates in image space, with model weights initialized from an unconditional diffusion model. In our experiments, we implement two variants of I$^2$SB for $4 \times$ super-resolution in latent space: one based on LDM and the other on AF-LDM. As shown in Fig.~\ref{fig:i2sb}, the standard latent I$^2$SB exhibits severe flickering and quality degradation when input shifts. In contrast, the alias-free variant demonstrates strong shift-equivariance, providing consistent outputs.

\noindent \textbf{Normal Estimation with YOSO}. Normal Estimation is another image-to-image task that requires the model to produce consistent outputs across shifts. We implement a one-step conditional diffusion model \cite{genpercept}, You-Only-Sample-Once (YOSO), following StableNormal \cite{stablenormal}. YOSO leverages ControlNet \cite{controlnet} to adapt a pre-trained SD into a normal map prediction model conditioned on input RGB images. For our implementation, we use $x_0$ parameterization instead of $x_t$ parameterization used in \cite{stablenormal}. We first train a baseline YOSO initialized from SD. We then enhance the ControlNet of YOSO with our anti-alias designs and train an alias-free YOSO (AF-YOSO) based on AF-SD. 

\begin{figure}[t]
\centering
\includegraphics[width=1.0\linewidth]{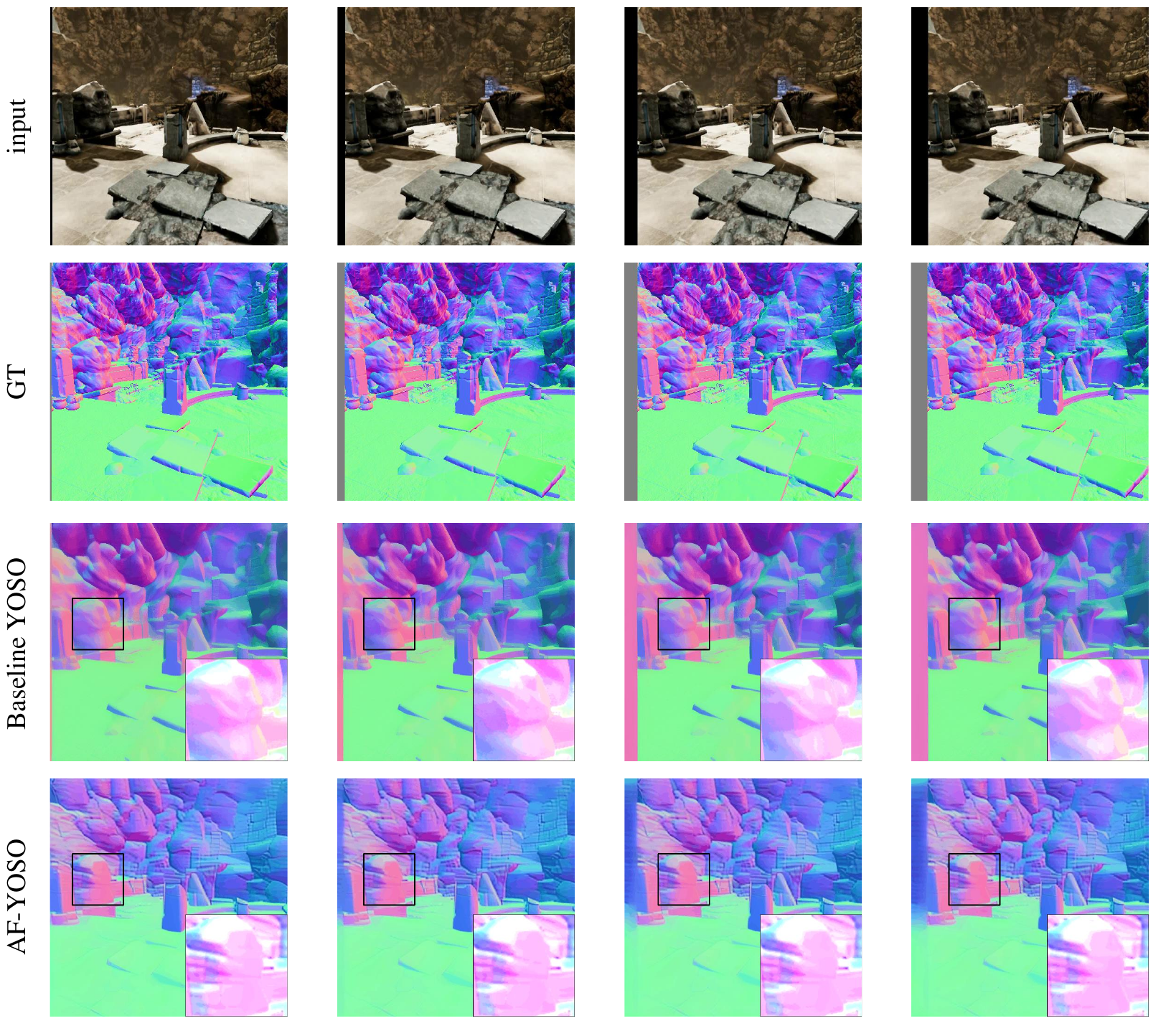}\vspace{-4mm}
\caption{Qualitative comparison of shift-equivariance in YOSO normal estimation. The contrast of local regions is enhanced. Please refer to the supplementary for a clearer video comparison.}\vspace{-4mm}
\label{fig:normal}
\end{figure}

We compare the shift-equivariance of vanilla YOSO and AF-YOSO in Fig.~\ref{fig:normal}. Although the baseline model produces largely consistent outputs, some flickering remains, particularly around high-frequent textures. In contrast, AF-YOSO delivers fully consistent results across the entire image, effectively mitigating flickering.

\section{Conclusion}
We present Alias-Free Latent Diffusion Models, a novel framework designed to eliminate the instability of LDMs when input shifts occur. By incorporating equivariance loss and equivariant attention along with anti-aliasing modules, our model achieves significantly improved shift-equivariance. Experimental results demonstrate that our model can facilitate various editing tasks, such as inversion-based video editing, providing consistent and stable outputs. Additionally, the approach can be extended to other tasks that leverage a pre-trained latent diffusion model, such as super-resolution and normal estimation, substantially enhancing consistency over spatial shifts. 

\small{
\noindent
\textbf{\diff{Acknowledgments.}} This research is supported by NTU SUG-NAP. This study is also supported under the RIE2020 Industry Alignment Fund – Industry Collaboration Projects (IAF-ICP) Funding Initiative, as well as cash and in-kind contribution from the industry partner(s).
}

{
    \small
    \bibliographystyle{ieeenat_fullname}
    \bibliography{main}
}

\clearpage

\setcounter{section}{0}
\renewcommand\thesection{\Alph{section}}

\section{Implementation Details}

\subsection{Alias-free Modules}

\diff{To obtain the fractional shift of a function $F$ applied to discretely sampled input $Z$, a common approach is to convert the $Z$ into ca ontinuous signal $z$, apply the fractional shift $T$ and corresponding function $f$ in the continuous domain, and convert the result $f(T(z))$ back to the discrete domain. Formally, we define }

\begin{equation}
    F(T(Z))= \text{to\_discrete}(f(T(z))),
\end{equation}
\diff{where $z=\text{to\_continuous(Z)}$.}

\diff{To ensure $F$ is shift-equivarint, we require}

\begin{equation}
\begin{split}
    F(T(Z)) &= \text{to\_discrete}(f(T(z))) \\
    &= \text{to\_discrete}(T(f(z))) \\
    &= T(F(Z)).
\end{split}    
\end{equation}
\diff{As noted in StyleGAN3~\cite{stylegan3}, the above equation holds when the conversions between $z$ and $Z$ are invertible, which in turn requires the signal $z$ to satisfy the sampling theorem~\cite{sampling_theorem}: the bandwidth of the signal $z$ must be smaller than half the sampling rate of $Z$. However, StyleGAN3 identifies two operations in common CNNs that violate this requirement}:
\begin{enumerate}
\item Nonlinearities, which introduce new high-frequency components into $z$.
\item Upsampling and downsampling, which may fail to correctly adjust the frequency of $z$.
\end{enumerate}

\diff{To address these issues}, we build upon prior work on alias-free neural networks \cite{af-convnet, stylegan3} and apply the following modifications to the network layers in both the VAE and U-Net architectures:

\noindent \textbf{Downsampling}. Strided downsampling convolutions are replaced with a standard convolution followed by a low-pass filter and nearest downsampling. The low-pass filter is implemented as an "ideal low-pass filter," removing high frequencies in the Fourier domain, as described in \cite{af-convnet}.

\noindent \textbf{Upsampling}. Upsampling is performed by zero-padding between existing pixels followed by convolution with a sinc interpolation kernel. Similar to the ``ideal low-pass filter" in downsampling, this convolution is executed in the Fourier domain via multiplication.

\noindent \textbf{Nonlinearity}. All nonlinearities, except those in the last layer, are wrapped between a $2 \times$ ideal upsampling and a $2 \times$ ideal downsampling. 

\subsection{VAE}

\begin{table} []
\caption{Hyperparameters of AF-VAE, AF-LDM, and AF-SD. All models are trained on 8 A100 GPUs.}\vspace{-2mm}
\label{tab:hyperparameters}
\resizebox{\linewidth}{!}{
\centering
\begin{tabular}{l|c|c|c}
\toprule
Model & (a) AF-VAE  & (b) AF-LDM & (c) AF-SD\\
\midrule 
Task & Reconstruction & Unconditional Generation & Text-to-Image \\ 
Dataset &  ImageNet &  FFHQ  & LAION Aesthetic 6.5+ \\
\midrule 
Resolution & $256 \times 256$ & $256 \times 256$ & $512 \times 512$ \\
Latent channels & 4 & 4 & 4\\
U-Net channels & 128, 256, 512, 512 & 192, 384, 384, 768, 768&320, 640, 1280, 1280\\ 
Attention Layers & - & 1 2 3 4 & 1 2 3 \\
Mid Block Attention & True & True & True \\
Head Channels & 512 & 24 & 8\\
Depth & 2 & 2 & 2 \\
Batch Size & 32 & 96 & 32 \\
Learning Rate & $5 \times 10^{-5}$ & $1 \times 10^{-4}$ & $1 \times 10^{-6}$\\
\bottomrule
\end{tabular}}\vspace{-2mm}
\end{table}

AF-VAE is initialized from the Stable Diffusion (SD) VAE (kl-f8 AE in \cite{ldm}). Alias-free modules replace the corresponding components in both the VAE and GAN discriminator \cite{vqgan}. Hyperparameters are detailed in Table~\ref{tab:hyperparameters}(a). The model is retrained on ImageNet using a combination of reconstruction, KL, GAN \cite{vqgan}, and equivariance loss.

\begin{equation}
L^{\text{VAE}} = L_{\text{rec}} + \lambda_1 L_{\text{KL}} + \lambda_2 L_{\text{GAN}} + \lambda_3 L_{\text{eq}}^{\text{VAE}}.
\end{equation}
We set $\lambda_1 = 10^{-6}, \lambda_2 = 0.25, \lambda_3 = 1$ by default. The equivariance loss of VAE is formulated as follows:

\begin{equation}
\begin{split}
    L_{\text{eq}}^{\text{VAE}} = \mathbb{E}_{x} (|| \left[ \mathcal{E}(T_{\Delta}(x)) - T_{\Delta / k}(\mathcal{E}(x)) \right] \cdot M_{\Delta / k} ||_2^2 + \\ || \left[ \mathcal{D}(T_{\Delta / k}(z)) - T_{\Delta  }(\mathcal{D}(z)) \right] \cdot M_{\Delta} ||_2^2),
\end{split}
\end{equation}
where $x \in \mathbb{R}^{H \times W \times 3}$ is the input image, $z = sg(\mathcal{E}(x)), z \in \mathbb{R}^{H/k \times W/k \times 4}$ is the latent downsampled by $k \times$, $sg$ is stop gradient operator, and $M_{\Delta}$ denotes the valid mask for cropped shift $T_{\Delta}$. Offsets $\Delta=(\Delta_x, \Delta_y) \in \mathbb{N}^2$. $\Delta_x, \Delta_y$ are uniformly sampled from $[-\frac{3}{8} H, \frac{3}{8} H]$ and $[-\frac{3}{8} W, \frac{3}{8} W]$, respectively. 

\subsection{Unconditional LDM}

After we obtain the AF-VAE, we train unconditional Latent Diffusion Models (LDMs) from scratch in its latent space. Similar to AF-VAE, alias-free modules are incorporated into the U-Net. The hyperparameters of U-Net are detailed in Table~\ref{tab:hyperparameters}(b). The training loss combines diffusion loss and equivariance loss.

\begin{equation}
L^{\text{LDM}} = \mathbb{E}_{\mathcal{E}(x), \epsilon \sim \mathcal{N}(0, 1), t}, [||\epsilon - \epsilon_{\theta}(z_t, t)||_2^2] + \lambda L_{\text{eq}}^{\text{LDM}},
\end{equation}
where $z_t$ is the noisy latent at timestep $t$. $\lambda$ is set to $1$ by default. The equivariance loss of LDM is defined as follows:

\begin{equation}
    L_{\text{eq}}^{\text{LDM}} = \mathbb{E}_{\mathcal{E}(x), t}|| \left[ \epsilon'_{\theta}(T_{\Delta}^{\text{cir}}(z_t), t) - T_{\Delta}(\epsilon_{\theta}(z_t, t)) \right] \cdot M_{\Delta} ||_2^2,
\end{equation}
where $\Delta=(\Delta_x / k, \Delta_y / k)$, $\Delta_x$ and $\Delta_y$ are sampled in the same way as VAE. $\epsilon'_{\theta}$ is the U-Net with equivariant attention.

\subsection{Text-conditional LDM}

We also train an alias-free text-conditional LDM (AF-SD) in the latent space of AF-VAE. Specifically, we initialize the U-Net from Stable Diffusion V1.5 \cite{ldm}, modify it with alias-free modules, and retrain it on the LAION Aesthetic 6.5+ dataset (Table~\ref{tab:hyperparameters}(c)). The loss function is similar to that of AF-LDM:

\begin{equation}
L^{\text{SD}} = \mathbb{E}_{\mathcal{E}(x), \epsilon \sim \mathbf{N}(0, 1), t, c}, [||\epsilon - \epsilon_{\theta}(z_t, t, c)||_2^2] + \lambda L_{\text{eq}}^{\text{LDM}},
\end{equation}
where $c$ is the text embedding of $x$ obtained from CLIP \cite{clip}.

\section{Details of Video Editing}

\setlength{\textfloatsep}{8pt}
\begin{algorithm}[t]
  \caption{Warping-equivariant Video Editing}
  \textbf{Input:} Video $X=\{x^i\}^N_{i=1}$, encoder $\mathcal{E}$, decoder $\mathcal{D}$, U-Net $\epsilon_{\theta}$, U-Net with cross-frame attention $\epsilon'_{\theta}$. \\
  \textbf{Output:} Edited video $\hat{X}=\{\hat{x}^i\}^N_{i=1}$.
  \begin{algorithmic}[1]
    \For{$i$ \textbf{in} [1, ..., N]}
        \State $z^i_0 \gets \mathcal{E}(x^i)$
    \EndFor
    \State $z^1_T \gets \texttt{DDIMInversion}(z^1_0, \epsilon_{\theta})$ \Comment{Cache K, V in $\epsilon'_{\theta}$}
    \For{$i$ \textbf{in} $[2, ..., N]$}
        \State $z^i_T \gets \texttt{DDIMInversion}(z^i_0, \epsilon'_{\theta})$
    \EndFor
    \State $\hat{z}^1_0 \gets \texttt{DDIMSampling}(z^1_T, \epsilon_{\theta})$ \Comment{Cache K, V in $\epsilon'_{\theta}$}
    \State $\hat{x}^1 \gets \mathcal{D}(\hat{z}^1_0)$
    \For{$i$ \textbf{in} $[2, ..., N]$}
        \State $\hat{z}^i_0 \gets \texttt{DDIMSampling}(z^i_0, \epsilon'_{\theta})$
        \State $\hat{x}^i \gets \mathcal{D}(\hat{z}^i_0)$
    \EndFor
  \end{algorithmic}
  \label{alg:algorithm1}
\end{algorithm}

\setlength{\textfloatsep}{8pt}
\begin{algorithm}[t]
  \caption{Image Splatting and Interpolation}
  \textbf{Input:} Input images $x^1, x^2$, encoder $\mathcal{E}$, decoder $\mathcal{D}$, U-Net $\epsilon_{\theta}$, U-Net with cross-frame attention to the first and second image $\epsilon'^1_{\theta}, \epsilon'^2_{\theta}$, interpolation frame numbers $N$. \\
  \textbf{Output:} Interpolation video $\hat{X}=\{\hat{x}^i\}^N_{i=1}$.
  \begin{algorithmic}[1]
    \State $f_{fwd}, f_{bwd} \gets \texttt{FlowEstimation}(x^1, x^2)$
    \State $z^1_0 \gets \mathcal{E}(x^1)$
    \State $z^2_0 \gets \mathcal{E}(x^2)$
    \State $z^1_T \gets \texttt{DDIMInversion}(z^1_0, \epsilon_{\theta})$
    \State $z^2_T \gets \texttt{DDIMInversion}(z^2_0, \epsilon_{\theta})$
    
    \State $\texttt{DDIMSampling}(z^1_0, \epsilon_{\theta})$ \Comment{Cache K, V in $\epsilon'^1_{\theta}$}
    \State $\texttt{DDIMSampling}(z^2_0, \epsilon_{\theta})$ \Comment{Cache K, V in $\epsilon'^2_{\theta}$}
    
    \For{$i$ \textbf{in} $[1, ..., N]$}
        \State $\alpha \gets 1 / (N+1)$
        \State $\epsilon'^\alpha_{\theta} \gets \texttt{KVInterpolation}(\epsilon'^1_{\theta}, \epsilon'^2_{\theta}, \alpha)$
        \State $z'^1_T \gets \texttt{Splatting}(z^1_T, \alpha \cdot f_{fwd})$
        \State $z'^2_T \gets \texttt{Splatting}(z^2_T, (1 - \alpha) \cdot f_{bwd})$
        \State $z^\alpha_T \gets \texttt{Slerp}(z'^1_T, z'^2_T, \alpha)$
        \State $\hat{z}^i_0 \gets \texttt{DDIMSampling}(z^\alpha_T, \epsilon'^\alpha_{\theta})$
        \State $\hat{x}^i \gets \mathcal{D}(\hat{z}^i_0)$
    \EndFor

  \end{algorithmic}
  \label{alg:algorithm2}
\end{algorithm}

\subsection{Warping-Equivariant Video Editing}

\begin{figure}[t]
\centering
\includegraphics[width=1.0\linewidth]{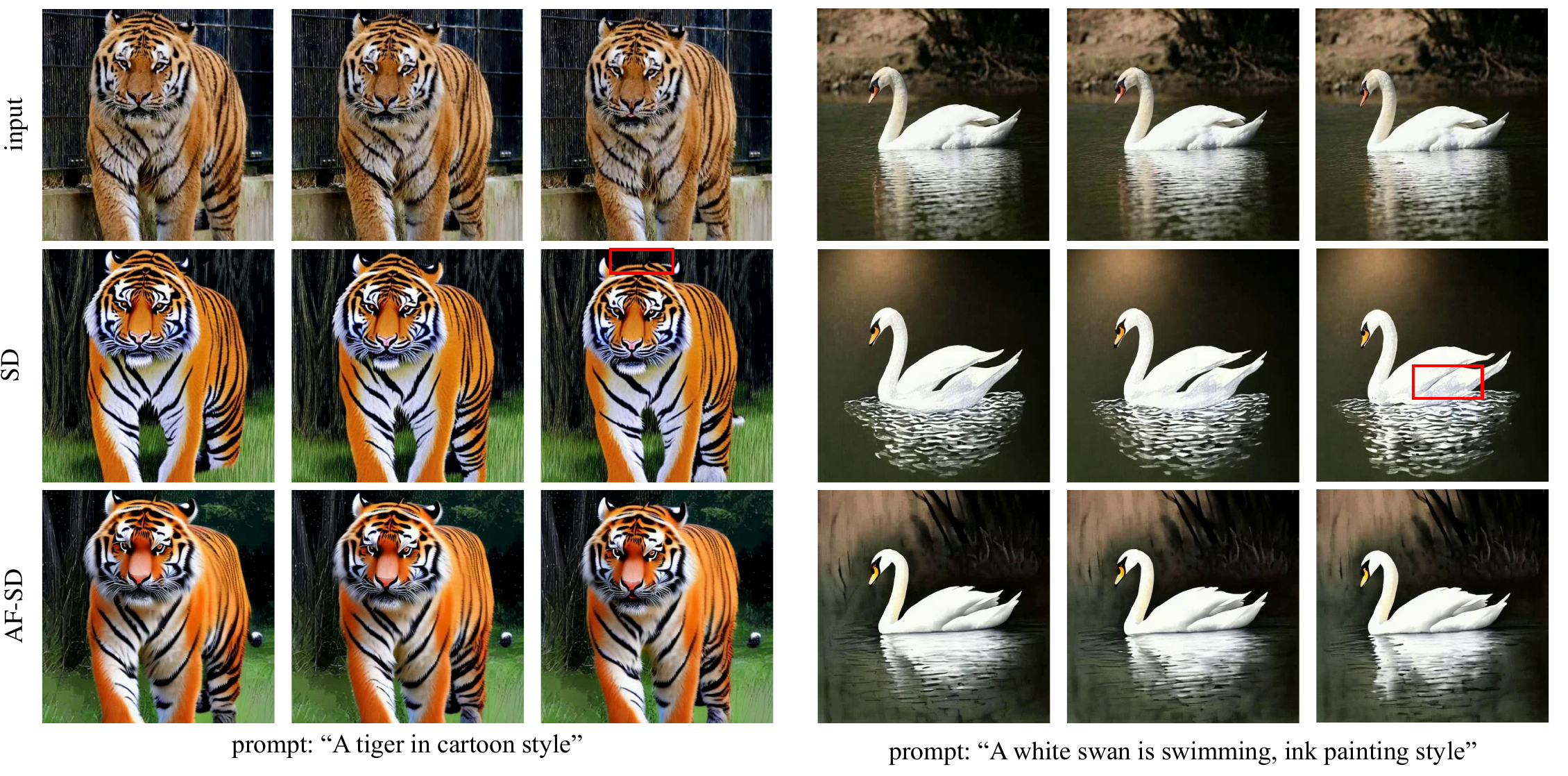}\vspace{-2mm}
\caption{Visualization of warping-equivariant video editing.}\vspace{-2mm}
\label{fig:video_edit_1}
\end{figure}

The algorithm of warping-equivariant video editing mentioned in the main text is illustrated in Alg.~\ref{alg:algorithm1}. Unlike standard inversion-based methods that invert each frame independently, we utilize cross-frame attention during inversion to naturally preserve deformation information in noisy latent. Additional results can be found in Fig.~\ref{fig:video_edit_1}.

\subsection{Image Splatting and Interpolation}

\begin{figure}[t]
\centering
\includegraphics[width=1.0\linewidth]{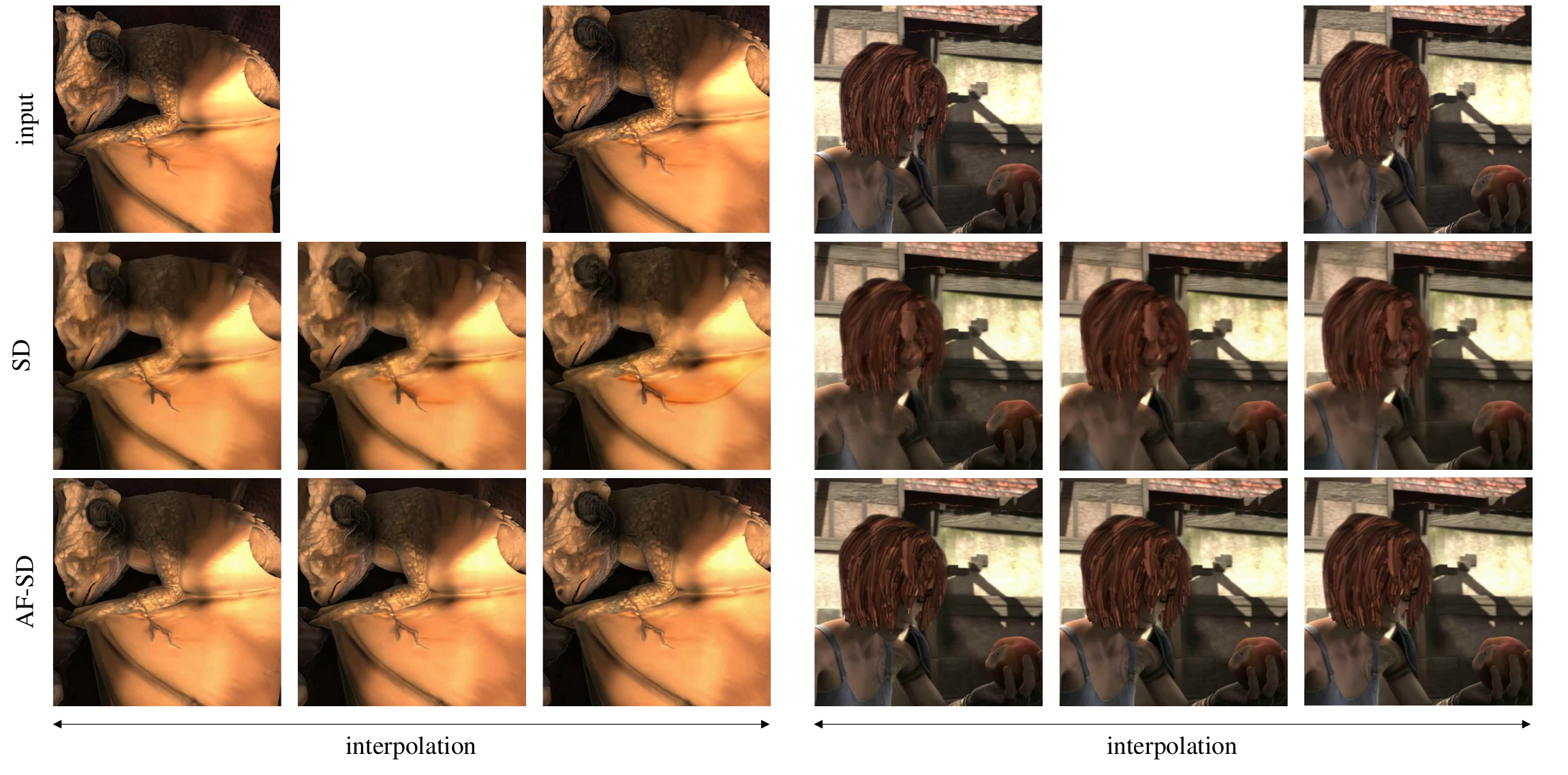}\vspace{-2mm}
\caption{Visualization of image interpolation. It is recommended 
to view it on the project page.}\vspace{-2mm}
\label{fig:video_edit_2}
\end{figure}

Leveraging AF-SD's high warping-equivariance, we also implement a simple image interpolation method using latent splatting (Alg.~\ref{alg:algorithm2}). Given two input images, we invert their latents, splat them using forward and backward flows with interpolation ratio $\alpha$, and interpolate the latents and attention features to generate intermediate frames \cite{diffmorpher, attn_interp}. We use GMFlow \cite{gmflow} as the flow estimator. Compared to standard SD, AF-SD produces smoother results. Although the method sometimes produces flickering artifacts in the occlusion region caused by latent splatting, this issue can be mitigated by integrating depth information and softmax splatting \cite{softmax_splatting}. To keep the algorithm as simple as possible, we do not apply this more advanced technique. The results are shown in Fig.~\ref{fig:video_edit_2}.

\section{\diff{Comparison to State-of-the-art Zero-shot Video Editing Methods}}

\begin{figure}[t]
\centering
\includegraphics[width=1.0\linewidth]{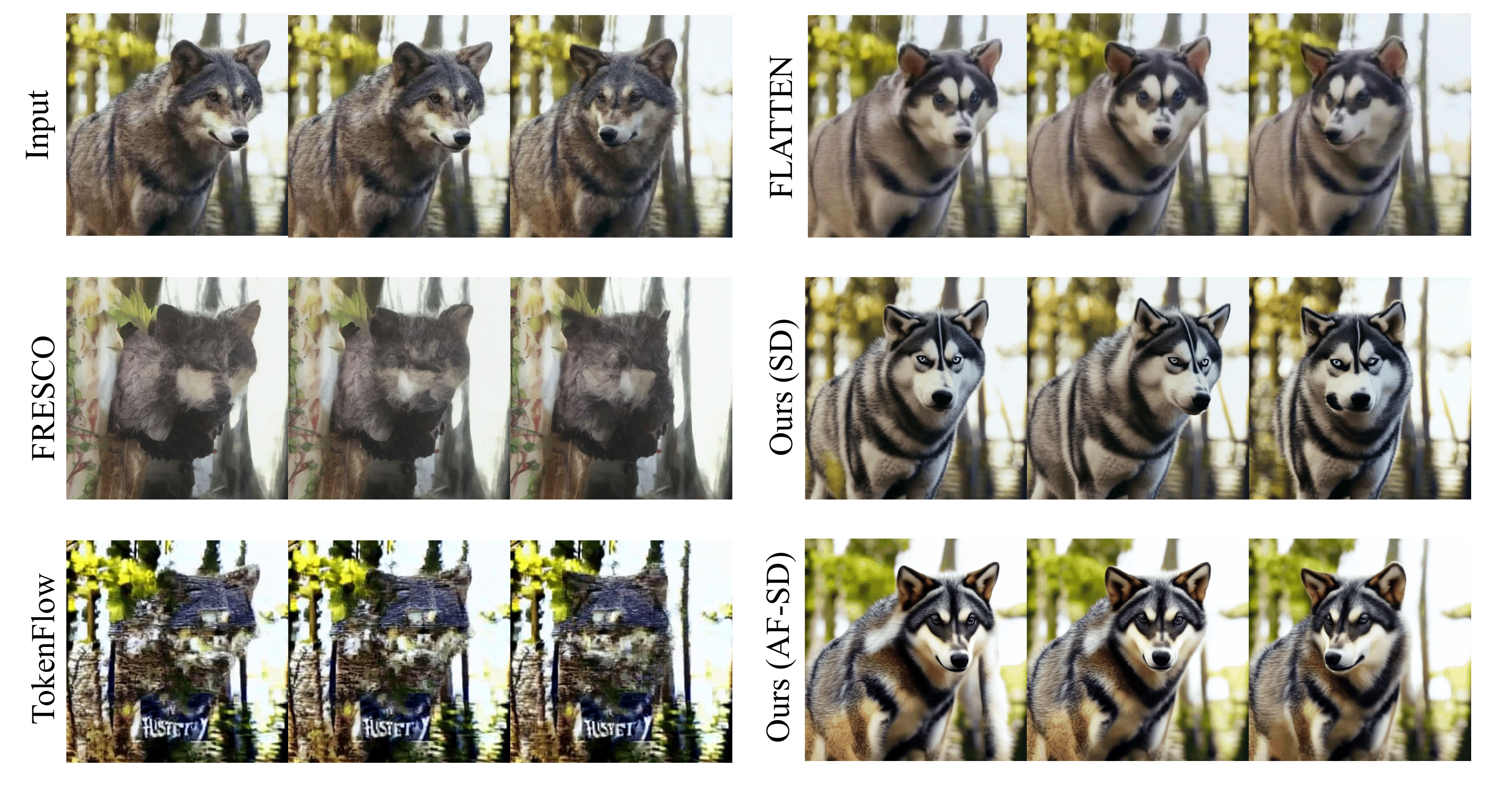}\vspace{-2mm}
\caption{{\footnotesize Comparison on video editing. Prompt: ``a wolf'' $\rightarrow$ ``A Husty''.}}\vspace{-3mm}
\label{fig:comparison}
\end{figure}

Figure~\ref{fig:comparison} and Table~\ref{tab:comparison} present both qualitative and quantitative comparisons with SOTA zero-shot editing methods. For a fair assessment, we use SD 1.5 as the backbone without enabling ControlNet. We conduct experiments on 13 video clips that exhibit small motions. Notably, FRESCO fails to produce satisfactory outputs without ControlNet. In contrast, our warping-equivariant editing pipeline (inversion and editing with cross-frame attention) with AF-SD achieves lower neighboring-frame warping error (measured as the MSE between the warped first frame and the second frame) than both TokenFlow and our SD version, highlighting the effectiveness of our alias-free design for consistent video editing. Although FLATTEN obtains an even lower warping error, it relies on an external flow estimation model, whereas our approach does not require any additional modules.

\begin{table} []
\caption{Quantitative comparison of video editing. }\vspace{-2mm}
\label{tab:comparison}
\resizebox{\linewidth}{!}{
\centering
\begin{tabular}{l|c|c|c|c|c}
\toprule
 & FRESCO & TokenFlow & FLATTEN & Ours (SD) & Ours (AF-SD)\\
\midrule 
Warping MSE $\downarrow$ & 0.325 & 0.064 & \textbf{0.036} & 0.094 & 0.056 \\
\bottomrule
\end{tabular}}
\vspace{-2mm}
\end{table}

\section{Limitation}

Our AF-LDM has a limitation: since we implement equivariant attention as cross-frame attention, all editing methods depend on a reference frame. This assumption means that objects or textures not present in the reference frame can lead to incoherent content in occlusion areas due to overlapping or new objects entering subsequent frames. Similar to flow-based video editing methods \cite{rerender, warp_noise}, our video editing method may produce incoherent results in static background regions where flow guidance is insufficient \cite{fresco}.

\section{More Qualitative Results}

\begin{figure*}[t]
\centering
\includegraphics[width=1.0\linewidth]{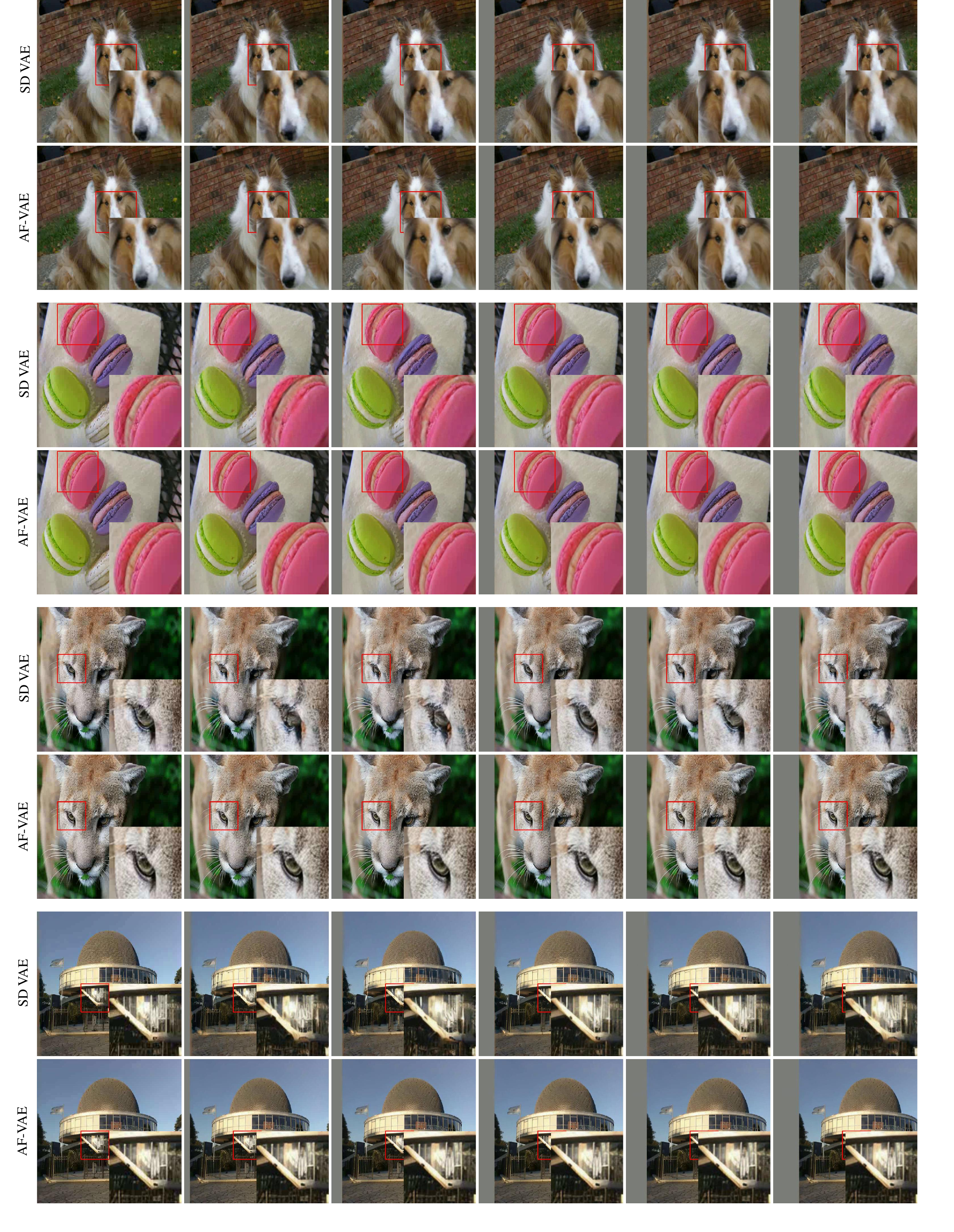}\vspace{-6mm}
\caption{Quantitative comparison of SD VAE and AF-VAE in latent shifting experiments. Shift offsets are 1, 19, 37, 55, 74, and 92 pixels.}\vspace{-2mm}
\label{fig:vae_supp}
\end{figure*}

\begin{figure*}[t]
\centering
\includegraphics[width=1.0\linewidth]{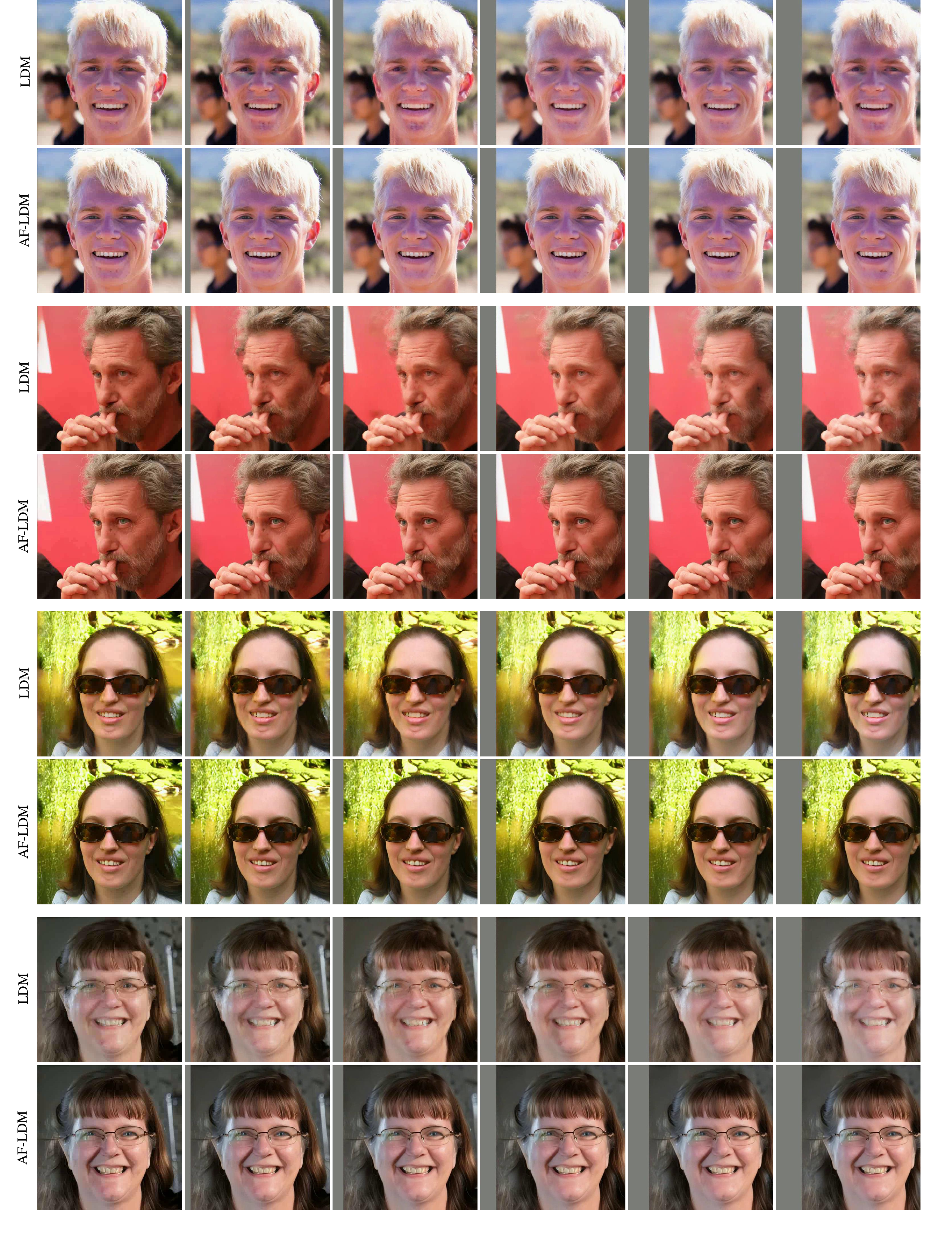}\vspace{-6mm}
\caption{Quantitative comparison of FFHQ unconditional LDM and AF-LDM in noisy latent shifting experiments. Latents are obtained from DDIM inversion. Shift offsets are 1, 19, 37, 55, 74, and 92 pixels. }\vspace{-2mm}
\label{fig:ldm_supp}
\end{figure*}

We present additional qualitative comparisons in this section, including AF-VAE vs. VAE (Fig~\ref{fig:vae_supp}) and AF-LDM vs. LDM (Fig~\ref{fig:ldm_supp}). It is recommended to watch the videos on the project page.

\end{document}